%% file: arxiv.tex
\documentclass{article}

\usepackage{microtype}
\usepackage{amsmath}
\usepackage{amssymb}
\usepackage{amsfonts}
\usepackage{graphicx}
\usepackage{subfigure}
\usepackage{booktabs} % for professional tables
\usepackage{url}
\usepackage{csquotes}
\usepackage{graphicx}
\usepackage{graphbox}
\usepackage{float}
\usepackage{multirow}
\usepackage{booktabs}

\usepackage{hyperref}

\usepackage[accepted]{icml2018}
\gdef\isarxiv{1}

\include{macros}

\icmltitlerunning{Synthesizing Robust Adversarial Examples}

\begin{document}

\twocolumn[
  \icmltitle{Synthesizing Robust Adversarial Examples}



\icmlsetsymbol{equal}{*}

\begin{icmlauthorlist}
\icmlauthor{Anish Athalye}{equal,mit,labsix}
\icmlauthor{Logan Engstrom}{equal,mit,labsix}
\icmlauthor{Andrew Ilyas}{equal,mit,labsix}
\icmlauthor{Kevin Kwok}{labsix}
\end{icmlauthorlist}

\icmlaffiliation{mit}{Massachusetts Institute of Technology}
\icmlaffiliation{labsix}{LabSix}

\icmlcorrespondingauthor{Anish Athalye}{aathalye@mit.edu}

\icmlkeywords{Machine Learning, ICML}

\vskip 0.3in
]



\printAffiliationsAndNotice{\icmlEqualContribution} % otherwise use the standard text.

\begin{abstract}
    \input{abstract}
\end{abstract}

\input{introduction}
\input{approach}
\input{evaluation}
\input{related-work}
\input{conclusion}
\input{acknowledgements}
\input{bibliography}

\clearpage
\appendix
\onecolumn
\input{appendix}

\end{document}

%% file: macros.tex
\usepackage{ifthen}
\usepackage{array}
\usepackage{xcolor}

\newcommand*{\todo}[2][]{\textcolor{red}{[\textbf{\ifthenelse{\equal{#1}{}}{TODO}{TODO(#1)}}: #2]}}

\DeclareMathOperator*{\argmax}{arg\,max}

\newcolumntype{L}[1]{>{\raggedright\let\newline\\\arraybackslash\hspace{0pt}}m{#1}}
\newcolumntype{C}[1]{>{\centering\let\newline\\\arraybackslash\hspace{0pt}}m{#1}}
\newcolumntype{R}[1]{>{\raggedleft\let\newline\\\arraybackslash\hspace{0pt}}m{#1}}

\newcommand{\legendbox}{\rule[-0.35ex]{2ex}{2ex}}
\newcommand{\clegendbox}[1]{\textcolor{#1}{\legendbox{}}}
\definecolor{legendgreen}{RGB}{39, 174, 96}
\definecolor{legendblack}{RGB}{0, 0, 0}
\definecolor{legendred}{RGB}{211, 47, 47}
\newcommand{\makelegend}[2]{
    \mbox{\clegendbox{legendgreen} classified as #1} \hspace{1em} \mbox{\clegendbox{legendred} classified as #2} \hspace{1em} \mbox{\clegendbox{legendblack} classified as other}
}

\newcommand*{\suppendix}{\ifdefined\isarxiv{appendix}\else{supplementary material}\fi}

\newcommand{\video}{
    \url{https://youtu.be/YXy6oX1iNoA}
}

%% file: abstract.tex
Standard methods for generating adversarial examples for neural networks do not
consistently fool neural network classifiers in the physical world due to a
combination of viewpoint shifts, camera noise, and other natural
transformations, limiting their relevance to real-world systems. We demonstrate
the existence of robust 3D adversarial objects, and we present the first
algorithm for synthesizing examples that are adversarial over a chosen
distribution of transformations. We synthesize two-dimensional adversarial
images that are robust to noise, distortion, and affine transformation. We
apply our algorithm to complex three-dimensional objects, using 3D-printing to
manufacture the first physical adversarial objects. Our results demonstrate the
existence of 3D adversarial objects in the physical world.

%% file: introduction.tex
\section{Introduction}
\label{sec:introduction}

The existence of adversarial examples for neural
networks~\cite{szegedy-intriguing,biggio2013evasion} was initially largely a
theoretical concern. Recent work has demonstrated the applicability of
adversarial examples in the physical world, showing that adversarial examples
on a printed page remain adversarial when captured using a cell phone camera in
an approximately axis-aligned setting~\cite{goodfellow-physical}. But while
minute, carefully-crafted perturbations can cause targeted misclassification in
neural networks, adversarial examples produced using standard techniques fail
to fool classifiers in the physical world when the examples are captured over
varying viewpoints and affected by natural phenomena such as lighting and
camera noise~\citep{luo-foveation,lu-noneed}. These results indicate that
real-world systems may not be at risk in practice because adversarial examples
generated using standard techniques are not robust in the physical world.

\begin{figure}[H]
	\begin{centering}
        \includegraphics[width=.8\linewidth]{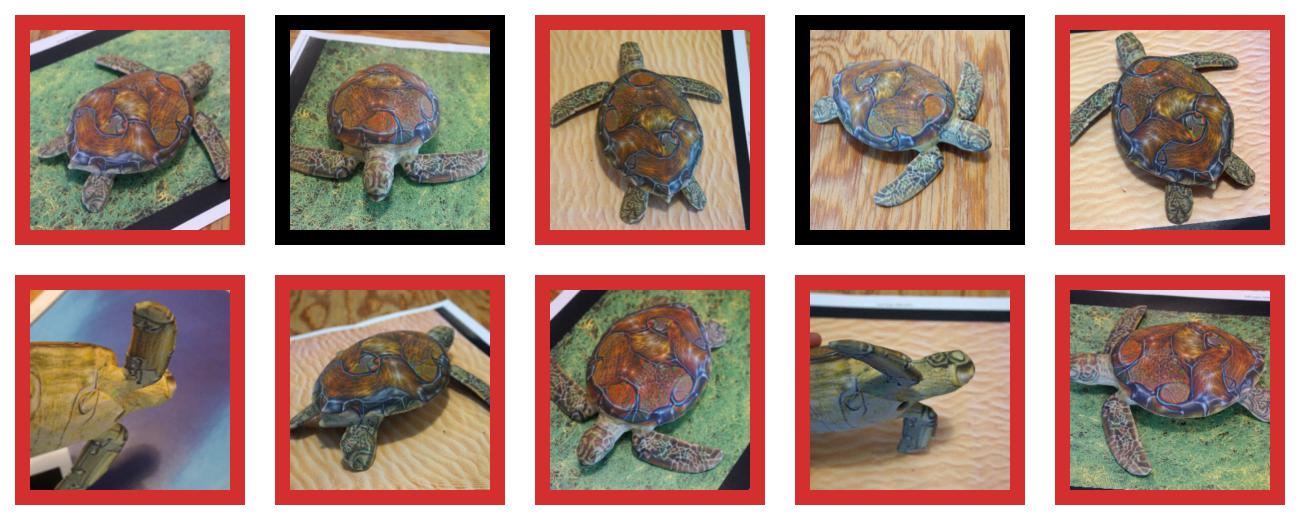}

        \makelegend{turtle}{rifle}
        \caption{
            Randomly sampled poses of a 3D-printed turtle adversarially
            perturbed to classify as a rifle at every
            viewpoint\protect\footnotemark{}. An unperturbed model is
            classified correctly as a turtle nearly 100\% of the time.
        }
        \label{fig:promoted-example}
    \end{centering}
\end{figure}

\footnotetext{See \video{} for a video where every
frame is fed through the ImageNet classifier: the turtle is consistently
classified as a rifle.}

We show that neural network-based classifiers are vulnerable to physical-world
adversarial examples that remain adversarial over a different viewpoints. We
introduce a new algorithm for synthesizing adversarial examples that are robust
over a chosen distribution of transformations, which we apply for reliably
producing robust adversarial images as well as physical-world adversarial
objects. Figure~\ref{fig:promoted-example} shows an example of an adversarial
object constructed using our approach, where a 3D-printed turtle is
consistently classified as rifle (a target class that was selected at random)
by an ImageNet classifier. In this paper, we demonstrate the efficacy and
generality of our method, demonstrating conclusively that adversarial examples
are a practical concern in real-world systems.

\subsection{Challenges}

Methods for transforming ordinary two-dimensional images into adversarial
examples, including techniques such as the L-BFGS
attack~\cite{szegedy-intriguing}, FGSM~\cite{iclr2015:goodfellow}, and the CW
attack~\cite{sp2017:carlini}, are well-known. While adversarial examples
generated through these techniques can transfer to the physical
world~\cite{goodfellow-physical}, the techniques have limited success in
affecting real-world systems where the input may be transformed before being
fed to the classifier. Prior work has shown that adversarial examples generated
using these standard techniques often lose their adversarial nature once subjected to
minor transformations~\cite{luo-foveation,lu-noneed}.

Prior techniques attempting to synthesize adversarial examples robust over any
chosen distribution of transformations in the physical world have had limited
success~\cite{evtimov-roadsigns}. While some progress has been made, concurrent
efforts have demonstrated a small number of data points on nonstandard
classifiers, and only in the two-dimensional case, with no clear generalization
to three dimensions (further discussed in Section~\ref{sec:related-work}).

Prior work has focused on generating two-dimensional adversarial examples, even
for the physical world~\cite{ccs2016:sharif,evtimov-roadsigns}, where
``viewpoints'' can be approximated by an affine transformations of an original
image. However, 3D objects must remain adversarial in the face of complex
transformations not applicable to 2D physical-world objects, such as 3D
rotation and perspective projection.

\subsection{Contributions}

We demonstrate the existence of robust adversarial examples and
adversarial objects in the physical world. We propose a general-purpose
algorithm for reliably constructing adversarial examples robust over a chosen
distribution of transformations, and we demonstrate the efficacy of this
algorithm in both the 2D and 3D case. We succeed in computing and fabricating
physical-world 3D adversarial objects that are robust over a large, realistic
distribution of 3D viewpoints, demonstrating that the algorithm successfully produces
adversarial three-dimensional objects that are adversarial in the physical
world. Specifically, our contributions are as follows:

\begin{itemize}

    \item We develop Expectation Over Transformation (EOT), the first algorithm that produces robust adversarial examples: single adversarial examples that are simultaneously adversarial over an entire distribution of transformations.

    \item We consider the problem of constructing 3D adversarial examples under the EOT framework, viewing the 3D rendering process as part of the transformation, and we show that the approach successfully synthesizes adversarial objects.

    \item We fabricate the first 3D physical-world adversarial objects and show that they fool classifiers in the physical world, demonstrating the efficacy of our approach end-to-end and showing the existence of robust physical-world adversarial objects.

\end{itemize}

%% file: approach.tex
\section{Approach}
\label{sec:approach}

First, we present the Expectation Over Transformation (EOT) algorithm, a
general framework allowing for the construction of adversarial examples that
remain adversarial over a chosen transformation distribution $T$. We then
describe our end-to-end approach for generating adversarial objects using a
specialized application of EOT in conjunction with differentiating through the
3D rendering process.

\subsection{Expectation Over Transformation}

When constructing adversarial examples in the white-box case (that is, with
access to a classifier and its gradient), we know in advance a set of possible
classes $Y$ and a space of valid inputs $X$ to the classifier; we have access to
the function
$P(y|x)$ and its gradient $\nabla_x P(y|x)$, for any class $y \in Y$ and
input $x \in X$. In the standard case, adversarial examples are produced by maximizing
the log-likelihood of the target class $y_t$ over a $\epsilon$-radius ball around the
original image (which we represent as a vector of $d$ pixels each in $[0,
1]$):

\begin{equation*}
\begin{aligned}
& \argmax_{x'}
& & \log P(y_t|x')
\\
& \text{subject to}
& & || x' - x ||_p < \epsilon  \\
&&& x' \in [0,1]^{d}
\end{aligned}
\end{equation*}

This approach has been shown to be effective at generating adversarial
examples. However, prior work has shown that these adversarial examples fail to
remain adversarial under image transformations that occur in the real world,
such as angle and viewpoint changes~\cite{luo-foveation,lu-noneed}.

To address
this issue, we introduce \textit{Expectation Over Transformation (EOT)}. The key
insight behind EOT is to model such perturbations within the optimization
procedure. Rather than optimizing the log-likelihood of a single
example, EOT uses a chosen distribution $T$ of transformation functions $t$
taking an input $x'$ controlled by the adversary to the ``true'' input $t(x')$
perceived by the classifier. Furthermore, rather than simply taking the norm of
$x' - x$ to constrain the solution space, given a distance function $d(\cdot,
\cdot)$, EOT instead aims to constrain the expected effective distance between
the adversarial and original inputs, which we define as:

$$\delta = \mathbb{E}_{t\sim T}[d(t(x'), t(x))]$$

We use this new definition because we want to minimize the (expected) perceived
distance as seen by the classifier. This is especially important in cases where
$t(x)$ has a different domain and codomain, e.g. when $x$ is a texture and
$t(x)$ is a rendering corresponding to the texture, we care to minimize the
visual difference between $t(x')$ and $t(x)$ rather than minimizing the
distance in texture space.

Thus, we have the following optimization problem:

\begin{equation*}
\begin{aligned}
& \argmax_{x'}
& & \mathbb{E}_{t \sim T}[\log P(y_t|t(x'))]
\\
& \text{subject to}
& & \mathbb{E}_{t \sim T}[d(t(x'), t(x))] < \epsilon \\
&&& x \in [0, 1]^d
\end{aligned}
\end{equation*}

In practice, the distribution $T$ can model perceptual distortions such as
random rotation, translation, or addition of noise. However, the method
generalizes beyond simple transformations; transformations in $T$ can perform
operations such as 3D rendering of a texture.

We maximize the objective via stochastic gradient descent. We
approximate the gradient of the expected value through sampling
transformations independently at each gradient descent step and differentiating
through the transformation.

\subsection{Choosing a distribution of transformations}

Given its ability to synthesize robust adversarial examples, we use the EOT
framework for generating 2D examples, 3D models, and ultimately physical-world
adversarial objects. Within the framework, however, there is a great deal of
freedom in the actual method by which examples are generated, including choice
of $T$, distance metric, and optimization method.

\subsubsection{2D case}

In the 2D case, we choose $T$ to approximate a realistic space of possible
distortions involved in printing out an image and taking a natural picture of
it. This amounts to a set of random transformations of the form $t(x) = Ax +
b$, which are more thoroughly described in Section~\ref{sec:evaluation}.
These random transformations are easy to differentiate, allowing for a
straightforward application of EOT.

\subsubsection{3D case}

We note that the domain and codomain of $t \in T$ need not be the same. To
synthesize 3D adversarial examples, we consider textures (color
patterns) $x$ corresponding to some chosen 3D object (shape), and we choose a
distribution of transformation functions $t(x)$ that take a texture and
render a pose of the 3D object with the texture $x$ applied. The
transformation functions map a texture to a rendering of an object, simulating
functions including rendering, lighting, rotation, translation, and perspective
projection of the object. Finding textures that are adversarial over a
realistic distribution of poses allows for transfer of adversarial examples to
the physical world.

To solve this optimization problem, EOT requires the ability to differentiate
though the 3D rendering function with respect to the texture. Given a
particular pose and choices for all other transformation parameters, a simple
3D rendering process can be modeled as a matrix multiplication and addition:
every pixel in the rendering is some linear combination of pixels in the
texture (plus some constant term). Given a particular choice of parameters, the
rendering of a texture $x$ can be written as $Mx + b$ for some
coordinate map $M$ and background $b$.

Standard 3D renderers, as part of the rendering pipeline, compute the
texture-space coordinates corresponding to on-screen coordinates; we modify an
existing renderer to return this information. Then, instead of differentiating
through the renderer, we compute and then differentiate through $Mx + b$. We
must re-compute $M$ and $b$ using the renderer for each pose, because EOT
samples new poses at each gradient descent step.

\subsection{Optimizing the objective}

Once EOT has been parameterized, i.e. once a distribution $T$ is chosen, the
issue of actually optimizing the induced objective function remains. Rather
than solving the constrained optimization problem given above, we use the
Lagrangian-relaxed form of the problem, as \citet{sp2017:carlini} do in the
standard single-viewpoint case:

\begin{equation*}
\begin{split}
    \argmax_{x'} \Bigl( \mathbb{E}_{t\sim T}[\log P(y_t|t(x'))] \\
    - \lambda \mathbb{E}_{t\sim T}[d(t(x'),t(x)]) \Bigr)
\end{split}
\end{equation*}

In order to encourage visual imperceptibility of the generated images, we set
$d(x',x)$ to be the $\ell_2$ norm in the LAB color space, a perceptually
uniform color space where Euclidean distance roughly corresponds with
perceptual distance~\cite{mclaren1976cielab}. Using distance in LAB space as a
proxy for human perceptual distance is a standard technique in computer vision.
Note that the $\mathbb{E}_{t\sim T}[||LAB(t(x')) - LAB(t(x))||_2]$ can be
sampled and estimated in conjunction with $\mathbb{E}[P(y_t|t(x))]$; in
general, the Lagrangian formulation gives EOT the ability to
constrain the search space (in our case, using LAB distance) without computing
a complex projection. Our optimization, then, is:

\begin{equation*}
\begin{split}
    \argmax_{x'} \mathbb{E}_{t\sim T}\Bigl[\log P(y_t|t(x')) \\
    - \lambda ||LAB(t(x')) -
LAB(t(x))||_2\Bigr]
\end{split}
\end{equation*}

We use projected gradient descent to maximize the objective, and clip to the set of
valid inputs (e.g. $[0, 1]$ for images).

%% file: evaluation.tex
\section{Evaluation}
\label{sec:evaluation}

First, we describe our procedure for quantitatively evaluating the efficacy of
EOT for generating 2D, 3D, and physical-world adversarial examples. Then, we
show that we can reliably produce transformation-tolerant adversarial examples
in both the 2D and 3D case. We show that we can synthesize and fabricate 3D
adversarial objects, even those with complex shapes, in the physical world:
these adversarial objects remain adversarial regardless of viewpoint, camera
noise, and other similar real-world factors. Finally, we present a qualitative
analysis of our results and discuss some challenges in applying EOT in the
physical world.

\subsection{Procedure}

In our experiments, we use TensorFlow's standard pre-trained InceptionV3
classifier~\citep{szegedy-inception} which has 78.0\% top-1 accuracy on
ImageNet. In all of our experiments, we use randomly chosen target classes, and
we use EOT to synthesize adversarial examples over a chosen distribution. We
measure the $\ell_2$ distance per pixel between the original and adversarial
example (in LAB space), and we also measure classification accuracy (percent of
randomly sampled viewpoints classified as the true class) and adversariality
(percent of randomly sampled viewpoints classified as the adversarial class)
for both the original and adversarial example. When working in simulation, we
evaluate over a large number of transformations sampled randomly from the
distribution; in the physical world, we evaluate over a large number of
manually-captured images of our adversarial objects taken over different
viewpoints.

Given a source object $x$, a set of correct classes $\{y_1, \ldots, y_n\}$, a
target class $y_{adv} \not\in \{y_1, \ldots, y_n\}$, and a robust adversarial
example $x'$, we quantify the effectiveness of the adversarial example over a
distribution of transformations $T$ as follows. Let $C(x, y)$ be a function
indicating whether the image $x$ was classified as the class $y$:

\[
C(x, y) = \begin{cases}
    1 & \text{if $x$ is classified as $y$} \\
    0 & \text{otherwise} \\
\end{cases}
\]

We quantify the effectiveness of a robust adversarial example by measuring
\textit{adversariality}, which we define as:

\[
    \mathbb{E}_{t \sim T}\left[ C(t(x'), y_{adv}) \right]
\]

This is equal to the probability that the example is classified as the target
class for a transformation sampled from the distribution $T$. We approximate
the expectation by sampling a large number of values from the distribution at
test time.

%We evaluate robust adversarial examples in the 2D and 3D case, and furthermore,
%we evaluate physical-world 3D adversarial objects. The two cases are
%fundamentally different. In the non-physical-world case, we know that we want
%to construct adversarial examples robust over a certain distribution of
%transformations, and we can simply use EOT over that distribution to synthesize
%a robust adversarial example. In the case of the physical world, however, we
%cannot capture the exact distribution unless we perfectly model all physical
%phenomena. Therefore, we must approximate the distribution and perform EOT over
%the proxy distribution. This works well in practice for producing adversarial
%objects that remain adversarial under the ``true'' physical-world distribution,
%as we demonstrate. See the supplementary material for the exact parameters of
%the distributions we use in the 2D, 3D simulation, and 3D physical-world cases.
%In all cases, the various transformation parameters are sampled as continuous
%random variables from a uniform distribution between the minimum and maximum
%values given, except for Gaussian noise.

\subsection{Robust 2D adversarial examples}

In the 2D case, we consider the distribution of transformations that includes rescaling, rotation, lightening or darkening by an additive factor, adding Gaussian noise, and translation of the image.

We take the first 1000 images in the ImageNet validation set, randomly choose
a target class for each image, and use EOT to synthesize an adversarial example
that is robust over the chosen distribution. We use a fixed $\lambda$ in our
Lagrangian to constrain visual similarity. For each adversarial example, we
evaluate over 1000 random transformations sampled from the distribution at
evaluation time. Table~\ref{tab:2d-results} summarizes the results. The
adversarial examples have a mean adversariality of 96.4\%, showing that our
approach is highly effective in producing robust adversarial examples.
Figure~\ref{fig:2d-examples} shows one synthesized adversarial example. See the
\suppendix{} for more examples.

% https://www.inf.ethz.ch/personal/markusp/teaching/guides/guide-tables.pdf
\begin{table*}
	\begin{centering}
        \begin{tabular}{c c c c c c c c c}
        \toprule
            \multirow{2}{*}{\textbf{Images}} & \phantom{x} & \multicolumn{2}{c}{\textbf{Classification Accuracy}} & \phantom{x} & \multicolumn{2}{c}{\textbf{Adversariality}} & \phantom{x} & \textbf{$\ell_2$} \\
            \cmidrule{3-4} \cmidrule{6-7} \cmidrule{9-9} && mean & stdev && mean & stdev && mean \\
        \midrule
            Original && 70.0\% & 36.4\% && 0.01\% & 0.3\% && 0 \\
            Adversarial && 0.9\% & 2.0\% && 96.4\%& 4.4\% && $5.6 \times 10^{-5}$ \\
        \bottomrule
        \end{tabular}
        \caption{
            Evaluation of $1000$ 2D adversarial examples with random targets. We evaluate each example over $1000$ randomly sampled transformations to calculate classification accuracy and adversariality (percent classified as the adversarial class).
        }
		\label{tab:2d-results}
    \end{centering}
\end{table*}

\begin{figure}
	\begin{centering}
        \input{./gen/2d-examples.tex}
        \caption{
        A 2D adversarial example showing classifier confidence in true /
        adversarial classes over randomly sampled poses.
        }
		\label{fig:2d-examples}
    \end{centering}
\end{figure}

\subsection{Robust 3D adversarial examples}
We produce 3D adversarial examples by modeling the 3D rendering as a transformation under EOT. Given a textured 3D object, we optimize the texture such that the rendering is adversarial from any viewpoint. We consider a distribution that incorporates different camera distances, lighting conditions, translation and rotation of the object, and solid background colors. We approximate the expectation over transformation by taking the mean loss over batches of size 40; furthermore, due to the computational expense of computing new poses, we reuse up to 80\% of the batch at each iteration, but enforce that each batch contain at least 8 new poses. As previously mentioned, the parameters of the distribution we use is specified in the \suppendix{}, sampled as independent continuous random variables (that are uniform except for Gaussian noise). We searched over several $\lambda$ values in our Lagrangian for each example / target class pair. In our final evaluation, we used the example with the smallest $\lambda$ that still maintained >90\% adversariality over 100 held out, random transformations.

We consider 10 3D models, obtained from 3D asset sites, that represent
different ImageNet classes: barrel, baseball, dog, orange, turtle, clownfish,
sofa, teddy bear, car, and taxi.

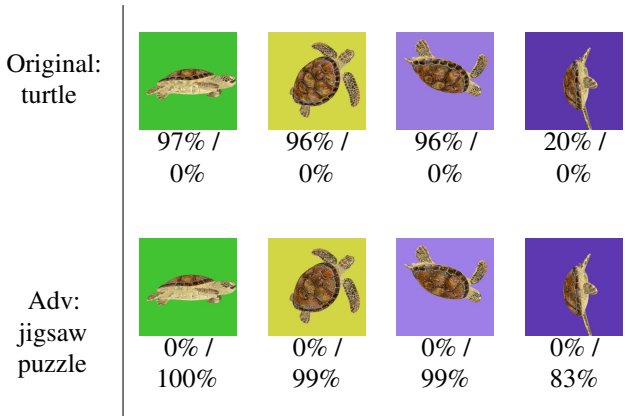
\begin{figure}
    \begin{centering}
        \input{./gen/3d-sim-examples.tex}
        \caption{
        A 3D adversarial example showing classifier confidence in true /
        adversarial classes over randomly sampled poses.
        }
        \label{fig:3d-sim-examples}
    \end{centering}
\end{figure}

We choose 20 random target classes per 3D model, and use EOT to
synthesize adversarial textures for the 3D models with minimal parameter
search (four pre-chosen $\lambda$ values were tested across each (3D model, target) pair). For each of the 200 adversarial examples, we sample 100 random transformations from the distribution at evaluation time. Table~\ref{tab:3d-sim-results} summarizes results, and Figure~\ref{fig:3d-sim-examples} shows renderings of drawn samples, along with classification probabilities. See the \suppendix{} for more examples.

The adversarial objects have a mean adversariality of 83.4\% with a long left tail, showing that EOT usually produces highly adversarial objects. See the \suppendix{} for a plot of the distribution of adversariality over the 200 examples.

\begin{table*}
	\begin{centering}
        \begin{tabular}{c c c c c c c c c}
        \toprule
            \multirow{2}{*}{\textbf{Images}} & \phantom{x} & \multicolumn{2}{c}{\textbf{Classification Accuracy}} & \phantom{x} & \multicolumn{2}{c}{\textbf{Adversariality}} & \phantom{x} & \textbf{$\ell_2$} \\
            \cmidrule{3-4} \cmidrule{6-7} \cmidrule{9-9} && mean & stdev && mean & stdev && mean \\
        \midrule
            Original && 68.8\% & 31.2\% && 0.01\% & 0.1\% && 0 \\
            Adversarial && 1.1\% & 3.1\% && 83.4\% & 21.7\% && $5.9 \times 10^{-3}$ \\
        \bottomrule
        \end{tabular}
        \caption{
            Evaluation of $200$ 3D adversarial examples with random targets. We
            evaluate each example over 100 randomly sampled poses to calculate
            classification accuracy and adversariality (percent classified as
            the adversarial class).
        }
        \label{tab:3d-sim-results}
    \end{centering}
\end{table*}

\subsection{Physical adversarial examples}

In the case of the physical world, we cannot capture the ``true'' distribution
unless we perfectly model all physical phenomena. Therefore, we must
approximate the distribution and perform EOT over the proxy distribution. We
find that this works well in practice: we produce objects that are optimized
for the proxy distribution, and we find that they generalize to the ``true''
physical-world distribution and remain adversarial.

Beyond modeling the 3D rendering process, we need to model physical-world
phenomena such as lighting effects and camera noise. Furthermore, we need to
model the 3D printing process: in our case, we use commercially available
full-color 3D printing. With the 3D printing technology we use, we find that
color accuracy varies between prints, so we model printing errors as well. We
approximate all of these phenomena by a distribution of transformations under
EOT: in addition to the transformations considered for 3D in simulation, we
consider camera noise, additive and multiplicative lighting, and per-channel
color inaccuracies.

\begin{figure}
	\begin{centering}
        \includegraphics[width=0.8\linewidth]{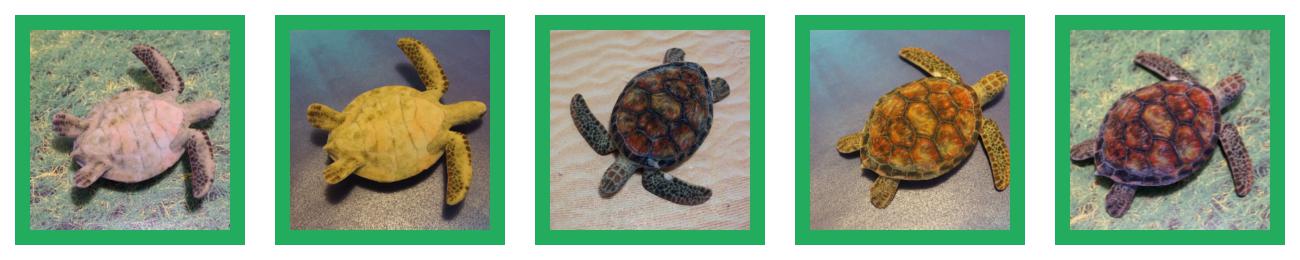}

        \makelegend{turtle}{rifle}

        \vspace{4ex}

        \includegraphics[width=0.8\linewidth]{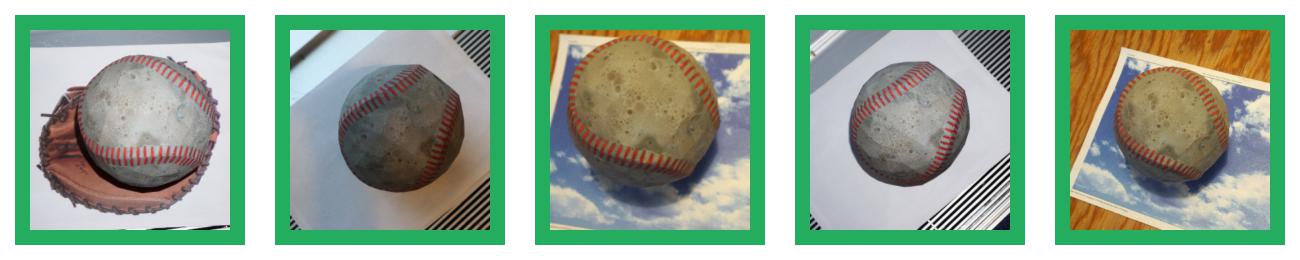}

        \makelegend{baseball}{espresso}
        \caption{
            A sample of photos of \textit{unperturbed} 3D prints. The
            unperturbed 3D-printed objects are consistently classified as the
            true class.
        }
        \label{fig:3d-originals}
    \end{centering}
\end{figure}

We evaluate physical adversarial examples over two 3D-printed objects: one of a turtle
(where we consider any of the 5 turtle classes in ImageNet as the ``true''
class), and one of a baseball. The unperturbed 3D-printed objects are correctly classified as
the true class with 100\% accuracy over a large number of samples.
Figure~\ref{fig:3d-originals} shows example photographs of unperturbed objects,
along with their classifications.

\begin{figure}
	\begin{centering}
        \includegraphics[width=\linewidth]{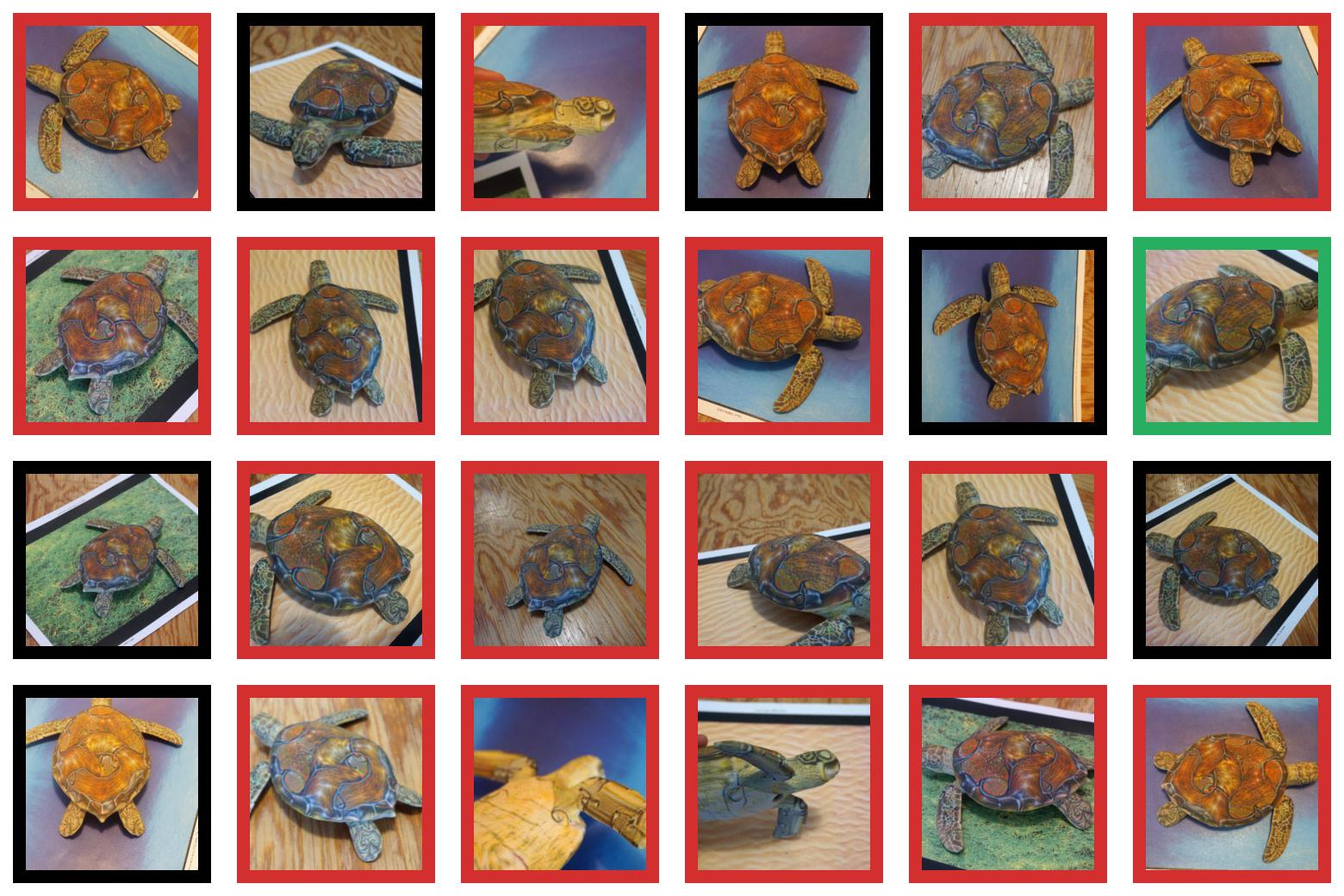}

        \makelegend{turtle}{rifle}

        \vspace{4ex}

        \includegraphics[width=\linewidth]{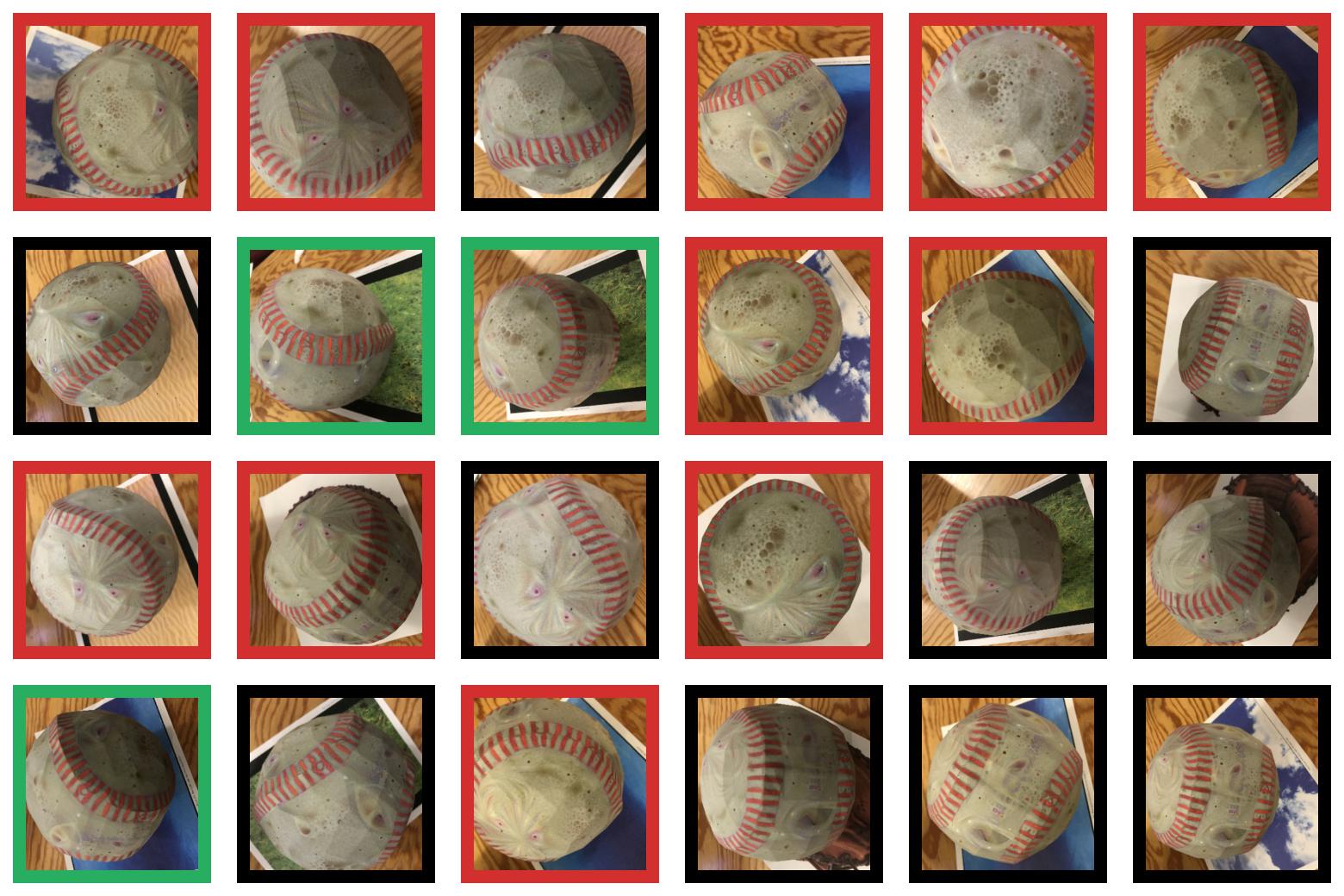}

        \makelegend{baseball}{espresso}
        \caption{
            Random sample of photographs of the \textit{two 3D-printed
            adversarial objects}. The 3D-printed adversarial objects are
            strongly adversarial over a wide distribution of viewpoints.
        }
        \label{fig:3d-examples}
    \end{centering}
\end{figure}

We choose target classes for each of the 3D models at random --- ``rifle'' for the
turtle, and ``espresso'' for the baseball --- and we use EOT to synthesize
adversarial examples. We evaluate the performance of our two 3D-printed
adversarial objects by taking 100 photos of each object over a variety of
viewpoints\footnote{Although the viewpoints were
simply the result of walking around the objects, moving them up/down, etc., we
do not call them ``random'' since they were not in fact generated
numerically or sampled from a concrete distribution, in contrast with the rendered
3D examples.}. Figure~\ref{fig:3d-examples} shows a random sample of these images,
along with their classifications. Table~\ref{tab:3d-results} gives a
quantitative analysis over all images, showing that our 3D-printed adversarial
objects are strongly adversarial over a wide distribution of transformations. See the \suppendix{} for more examples.

\begin{table}
	\begin{centering}
        \begin{tabular}{c c c c c}
        \toprule
            \textbf{Object} & \phantom{abc} & \textbf{Adversarial} & \textbf{Misclassified} & \textbf{Correct} \\
            \midrule
            Turtle && 82\% & 16\% & 2\% \\
            Baseball && 59\% & 31\% & 10\% \\
            \bottomrule
        \end{tabular}
        \caption{
            Quantitative analysis of the two adversarial objects, over 100
            photos of each object over a wide distribution of viewpoints. Both
            objects are classified as the adversarial target class in the
            majority of viewpoints.
        }
		\label{tab:3d-results}
    \end{centering}
\end{table}

\subsection{Discussion}

Our quantitative analysis demonstrates the efficacy of EOT and confirms the
existence of robust physical-world adversarial examples and objects. Now, we
present a qualitative analysis of the results.

\paragraph{Perturbation budget. } The perturbation required to produce
successful adversarial examples depends on the distribution of transformations
that is chosen. Generally, the larger the distribution, the larger the
perturbation required. For example, making an adversarial example robust to
rotation of up to $30^\circ$ requires less perturbation than making an example
robust to rotation, translation, and rescaling. Similarly, constructing robust
3D adversarial examples generally requires a larger perturbation to the
underlying texture than required for constructing 2D adversarial examples.

\begin{figure}
    \begin{centering}
	\includegraphics[width=0.2\linewidth]{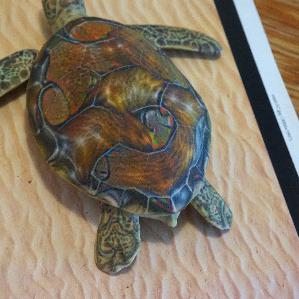}
	\includegraphics[width=0.2\linewidth]{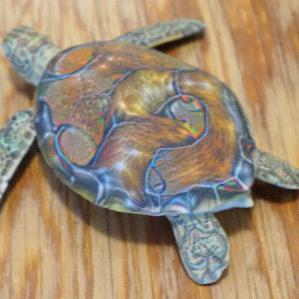}
	\includegraphics[width=0.2\linewidth]{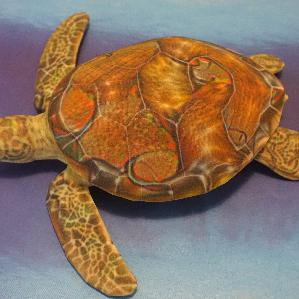}
    \caption{Three pictures of the same adversarial turtle (all classified as
        ``rifle''), demonstrating the need for a wide distribution and the
        efficacy of EOT in finding examples robust across wide distributions of
        physical-world effects like lighting.}
		\label{fig:orange_turtle}
    \end{centering}
\end{figure}

\paragraph{Modeling perception. } The EOT algorithm as presented in
Section~\ref{sec:approach} presents a general method to construct adversarial
examples over a chosen perceptual distribution, but notably gives no guarantees
for observations of the image outside of the chosen distribution. In
constructing physical-world adversarial objects, we use a crude approximation
of the rendering and capture process, and this succeeds in ensuring robustness
in a diverse set of environments; see, for example,
Figure~\ref{fig:orange_turtle}, which shows the same adversarial turtle in
vastly different lighting conditions. When a stronger guarantee is needed, a
domain expert may opt to model the perceptual distribution more precisely in
order to better constrain the search space.

\begin{figure}
    \begin{centering}
    	\includegraphics[width=.2\linewidth]{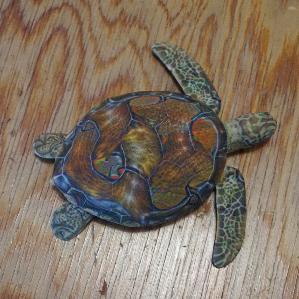}
    	\includegraphics[width=.2\linewidth]{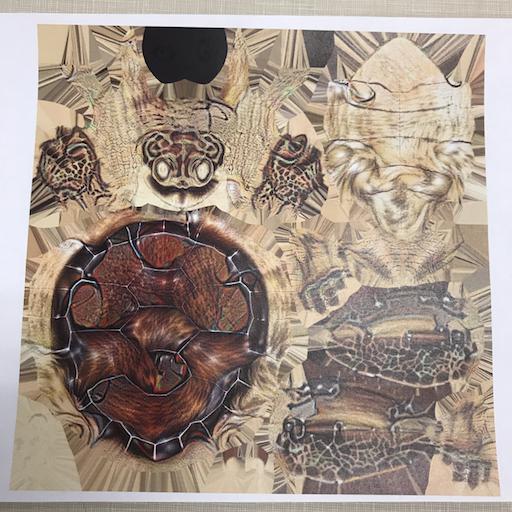}
    	\includegraphics[width=.2\linewidth]{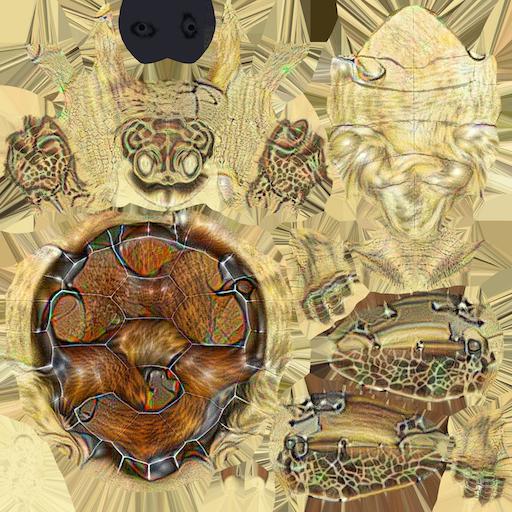}
        \caption{
	    A side-by-side comparison of a 3D-printed model (left) along with a
	    printout of the corresponding texture, printed on a standard laser
	    color printer (center) and the original digital texture (right),
	    showing significant error in color accuracy in printing.
        }
	\label{fig:color_acc}
    \end{centering}
\end{figure}

\paragraph{Error in printing. } We find significant error in the color accuracy
of even state of the art commercially available color 3D printing;
Figure~\ref{fig:color_acc} shows a comparison of a 3D-printed model along with
a printout of the model's texture, printed on a standard laser color printer.
Still, by modeling this color error as part of the distribution of
transformations in a coarse-grained manner, EOT was able to overcome the
problem and produce robust physical-world adversarial objects. We predict that
we could have produced adversarial examples with smaller $\ell_2$ perturbation
with a higher-fidelity printing process or a more fine-grained model
incorporating the printer's color gamut.

\paragraph{Semantically relevant misclassification. } Interestingly, for the
majority of viewpoints where the adversarial target class is not the top-1
predicted class, the classifier also fails to correctly predict the source
class. Instead, we find that the classifier often classifies the object as an
object that is \textit{semantically similar} to the adversarial target; while
generating the adversarial turtle to be classified as a rifle, for example, the
second most popular class (after ``rifle'') was ``revolver,'' followed by
``holster'' and then ``assault rifle.'' Similarly, when
generating the baseball to be classified as an espresso, the example was often
classified as ``coffee'' or ``bakery.''

\paragraph{Breaking defenses. } The existence of robust adversarial examples
implies that defenses based on randomly transforming the input are not secure:
adversarial examples generated using EOT can circumvent these defenses.
\citet{anish-carlini} investigates this further and circumvents several
published defenses by applying Expectation over Transformation.

\paragraph{Limitations. } There are two possible failure cases of the EOT algorithm.
As with any adversarial attack, if the attacker is constrained
to too small of a $\ell_p$ ball, EOT will be unable to create an adversarial example.
Another case is when the distribution of transformations
the attacker chooses is too ``large''.
As a simple example, it is impossible to make an adversarial example robust to
the function that randomly perturbs each pixel value to the interval $[0,1]$
uniformly at random.

\paragraph{Imperceptibility. } Note that we consider a ``targeted adversarial example'' to be an input that has been perturbed to misclassify as a selected class, is within the $\ell_p$ constraint bound imposed, and can be still clearly identified as the original class. While many of the generated examples are truly imperceptible from their corresponding original inputs, others exhibit noticeable perturbations. In all cases, however, the visual constraint ($\ell_2$ metric) maintains identifiability as the original class.

%% file: gen/2d-examples.tex
\begingroup\renewcommand*{\arraystretch}{3}
\begin{tabular}{C{0.17500\linewidth}|C{0.15625\linewidth}C{0.15625\linewidth}C{0.15625\linewidth}C{0.15625\linewidth}}
\includegraphics[align=c,width=\linewidth]{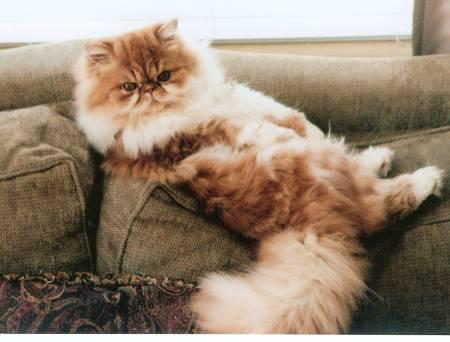} \newline Original: Persian cat & \includegraphics[align=c,width=\linewidth]{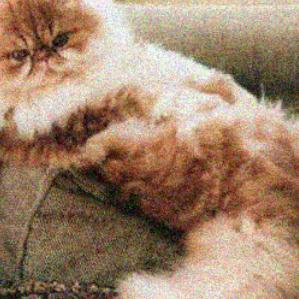} \newline 97\% / 0\% & \includegraphics[align=c,width=\linewidth]{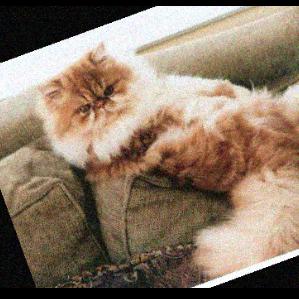} \newline 99\% / 0\% & \includegraphics[align=c,width=\linewidth]{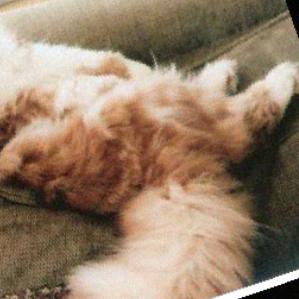} \newline 19\% / 0\% & \includegraphics[align=c,width=\linewidth]{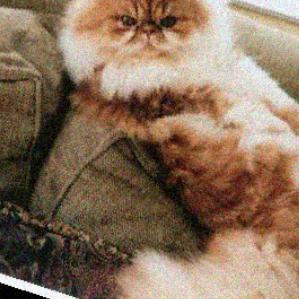} \newline 95\% / 0\% \\ 
\includegraphics[align=c,width=\linewidth]{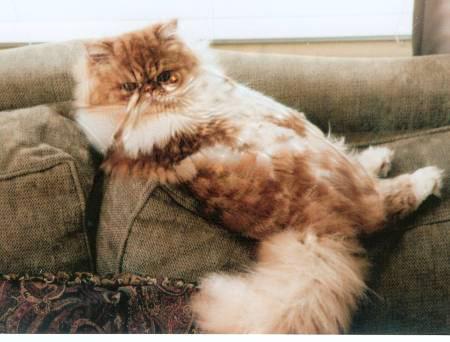} \newline Adv: jacamar& \includegraphics[align=c,width=\linewidth]{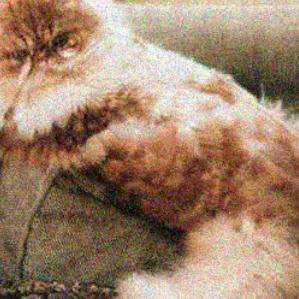} \newline 0\% / 91\% & \includegraphics[align=c,width=\linewidth]{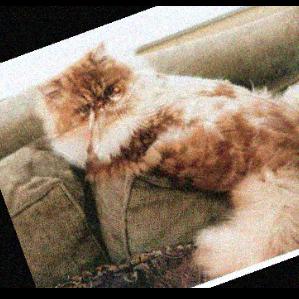} \newline 0\% / 96\% & \includegraphics[align=c,width=\linewidth]{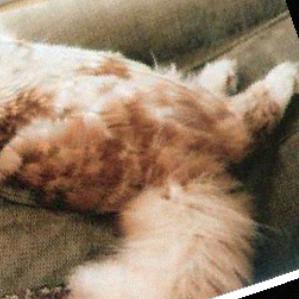} \newline 0\% / 83\% & \includegraphics[align=c,width=\linewidth]{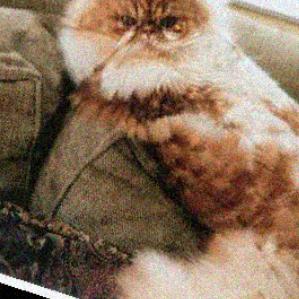} \newline 0\% / 97\% \\ 
\end{tabular}
\endgroup

%% file: gen/3d-sim-examples.tex
\begingroup\renewcommand*{\arraystretch}{3}
\begin{tabular}{C{0.17500\linewidth}|C{0.15625\linewidth}C{0.15625\linewidth}C{0.15625\linewidth}C{0.15625\linewidth}}
Original: turtle \vspace{8ex}& \includegraphics[align=c,width=\linewidth]{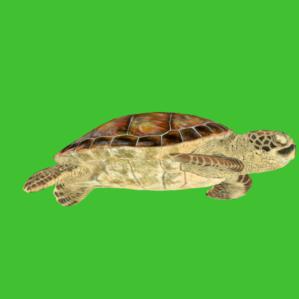} \newline 97\% / 0\% & \includegraphics[align=c,width=\linewidth]{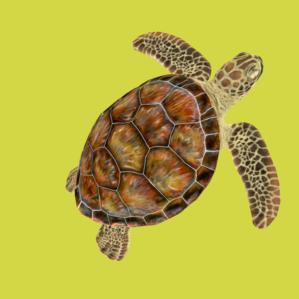} \newline 96\% / 0\% & \includegraphics[align=c,width=\linewidth]{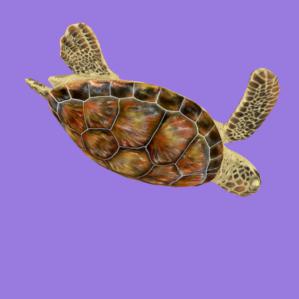} \newline 96\% / 0\% & \includegraphics[align=c,width=\linewidth]{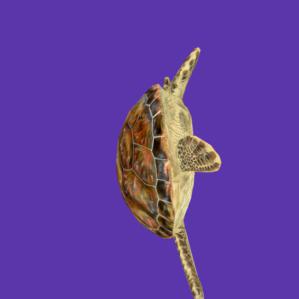} \newline 20\% / 0\% \\ 
Adv: jigsaw puzzle& \includegraphics[align=c,width=\linewidth]{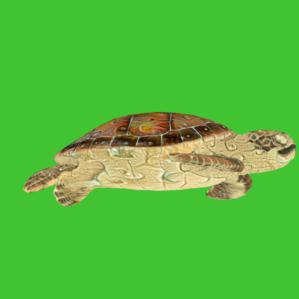} \newline 0\% / 100\% & \includegraphics[align=c,width=\linewidth]{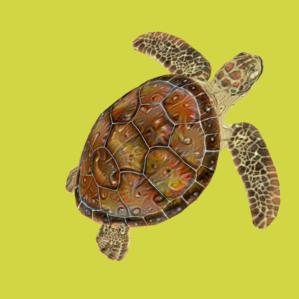} \newline 0\% / 99\% & \includegraphics[align=c,width=\linewidth]{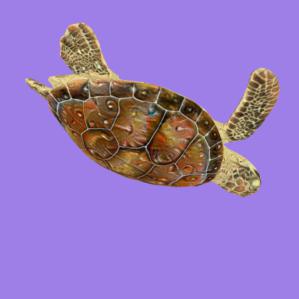} \newline 0\% / 99\% & \includegraphics[align=c,width=\linewidth]{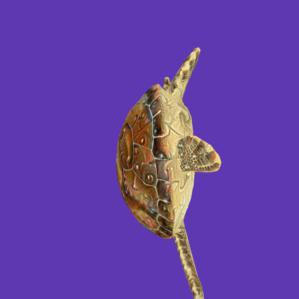} \newline 0\% / 83\% \\ 
\end{tabular}
\endgroup

%% file: related-work.tex
\section{Related Work}
\label{sec:related-work}

\subsection{Adversarial examples}

State of the art neural networks are vulnerable to adversarial
examples~\citep{szegedy-intriguing,biggio2013evasion}. Researchers have
proposed a number of methods for synthesizing adversarial examples in the
white-box setting (with access to the gradient of the classifier), including
L-BFGS~\citep{szegedy-intriguing}, the Fast Gradient Sign Method
(FGSM)~\citep{iclr2015:goodfellow}, Jacobian-based Saliency Map Attack
(JSMA)~\citep{sp2016:papernot}, a Lagrangian relaxation
formulation~\citep{sp2017:carlini}, and DeepFool~\citep{Moosavi-Dezfooli15},
all for what we call the single-viewpoint case where the adversary directly
controls the input to the neural network. Projected Gradient Descent (PGD) can
be seen as a universal first-order adversary~\citep{madry-adversarial}. A
number of approaches find adversarial examples in the black-box setting, with
some relying on the transferability phenomena and making use of substitute
models~\citep{papernot17,papernot-transferability} and others applying
black-box gradient estimation~\citep{zoo}.

\citet{cvpr2017:dezfooli} show the existence of universal (image-agnostic)
adversarial perturbations, small perturbation vectors that can be applied to
any image to induce misclassification. Their work solves a different problem than we do:
they propose an algorithm that finds perturbations that are universal over
images; in our work, we give an algorithm that finds a perturbation to a single
image or object that is universal over a chosen distribution of
transformations. In preliminary experiments, we found that universal
adversarial perturbations, like standard adversarial perturbations to single
images, are not inherently robust to transformation.

\subsection{Defenses}

Some progress has been made in defending against adversarial examples in the
white-box setting, but a complete solution has not yet been found. Many
proposed
defenses~\citep{papernot2016distillation,hendrik2017detecting,hendrycks2017early,meng2017magnet,zantedeschi2017efficient,buckman2018thermometer,ma2018characterizing,guo2018countering,dhillon2018stochastic,xie2018mitigating,song2018pixeldefend,samangouei2018defensegan}
have been found to be vulnerable to iterative optimization-based
attacks~\citep{carlini2016distillation,sp2017:carlini,carlini2017magnet,carlini2017adversarial,anish-carlini}.

Some of these defenses that can be viewed as ``input transformation'' defenses
are circumvented through application of EOT.

\subsection{Physical-world adversarial examples}

In the first work on physical-world adversarial examples,
\citet{goodfellow-physical} demonstrate the transferability of FGSM-generated
adversarial misclassification on a printed page. In their setup, a photo is
taken of a printed image with QR code guides, and the resultant image is
warped, cropped, and resized to become a square of the same size as the source
image before classifying it. Their results show the existence of 2D
physical-world adversarial examples for approximately axis-aligned views,
demonstrating that adversarial perturbations produced using FGSM can transfer
to the physical world and are robust to camera noise, rescaling, and lighting
effects. \citet{goodfellow-physical} do not synthesize targeted physical-world
adversarial examples, they do not evaluate other real-world 2D transformations
such as rotation, skew, translation, or zoom, and their approach does not
translate to the 3D case.

\citet{ccs2016:sharif} develop a real-world adversarial attack on a
state-of-the-art face recognition algorithm, where adversarial eyeglass frames
cause targeted misclassification in portrait photos. The algorithm produces
robust perturbations through optimizing over a fixed set of inputs: the
attacker collects a set of images and finds a perturbation that minimizes cross
entropy loss over the set. The algorithm solves a different problem than we do
in our work: it produces adversarial perturbations universal over portrait
photos taken head-on from a single viewpoint, while EOT produces 2D/3D
adversarial examples robust over transformations. Their approach also includes a
mechanism for enhancing perturbations' printability using a color map to
address the limited color gamut and color inaccuracy of the printer. Note that
this differs from our approach in achieving printability: rather than creating
a color map, we find an adversarial example that is robust to color inaccuracy.
Our approach has the advantage of working in settings where color accuracy
varies between prints, as was the case with our 3D-printer.

Concurrently to our work, \citet{evtimov-roadsigns} proposed a method for generating robust
physical-world adversarial examples in the 2D case by optimizing over a fixed
set of manually-captured images. However, the approach is limited to the 2D
case, with no clear translation to 3D, where there is no simple mapping between
what the adversary controls (the texture) and the observed input to the
classifier (an image). Furthermore, the approach requires the taking and
preprocessing of a large number of photos in order to produce each adversarial
example, which may be expensive or even infeasible for many objects.

\citet{brown2017patch} apply our EOT algorithm to produce an ``adversarial
patch'', a small image patch that can be applied to any scene to cause targeted
misclassification in the physical world.

Real-world adversarial examples have also been demonstrated in contexts other
than image classification/detection, such as
speech-to-text~\cite{carlini2016hidden}.

%% file: conclusion.tex
\section{Conclusion}
\label{sec:conclusion}

Our work demonstrates the existence of robust adversarial examples, adversarial
inputs that remain adversarial over a chosen distribution of transformations.
By introducing EOT, a general-purpose algorithm for creating robust adversarial
examples, and by modeling 3D rendering and printing within the framework of
EOT, we succeed in fabricating three-dimensional adversarial objects. With
access only to low-cost commercially available 3D printing technology, we
successfully print physical adversarial objects that are classified as a chosen
target class over a variety of angles, viewpoints, and lighting conditions by a
standard ImageNet classifier. Our results suggest that adversarial examples and
objects are a practical concern for real world systems, even when the examples
are viewed from a variety of angles and viewpoints.

\ifdefined\isarxiv
\else
\ifnum\value{page}>8
\todo{paper is too long: limit is 8 pages for main content}
\else
\fi
\fi

%% file: acknowledgements.tex
\clearpage
\section*{Acknowledgments}

We wish to thank Ilya Sutskever for providing feedback on early parts of this
work, and we wish to thank John Carrington and ZVerse for providing financial
and technical support with 3D printing. We are grateful to Tatsu Hashimoto,
Daniel Kang, Jacob Steinhardt, and Aditi Raghunathan for helpful comments on early drafts of this paper.

%% file: bibliography.tex
\bibliography{paper}
\bibliographystyle{icml2018}

%% file: appendix.tex
\section{Distributions of Transformations}
\label{app:distribution}

Under the EOT framework, we must choose a distribution of transformations, and
the optimization produces an adversarial example that is robust under the
distribution of transformations. Here, we give the specific parameters we chose
in the 2D (Table~\ref{tab:2d-dist}), 3D (Table~\ref{tab:3d-sim-dist}), and physical-world case (Table~\ref{tab:3d-dist}).

\section{Robust 2D Adversarial Examples}
\label{app:2d}

We give a random sample out of our 1000 2D adversarial examples in Figures~\ref{fig:2d-appendix-1}~and~\ref{fig:2d-appendix-2}.

\section{Robust 3D Adversarial Examples}
\label{app:3d-sim}

\begin{figure}[H]
      \begin{centering}
          \includegraphics[width=.9\linewidth]{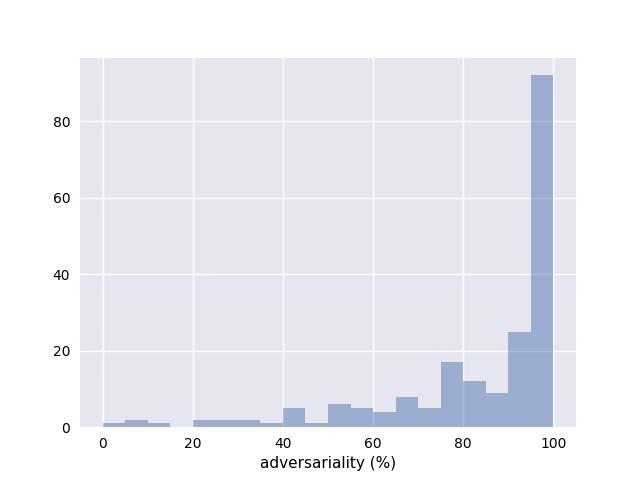}
          \caption{
              A histogram of adversariality (percent of 100 samples classified as the adversarial class) across the 200 3D adversarial examples.
          }
          \label{fig:3d-ex-hist}
      \end{centering}
  \end{figure}

We give a random sample out of our 200 3D adversarial examples in
Figures~\ref{fig:3d-sim-appendix-1}~and~\ref{fig:3d-sim-appendix-2}~and~\ref{fig:3d-sim-appendix-3}.
We give a histogram of adversariality (percent classified as the adversarial
class) over all 200 examples in Figure~\ref{fig:3d-ex-hist}.

\section{Physical Adversarial Examples}
\label{app:3d}

\begin{figure}[H]
	\begin{centering}
        \includegraphics[width=\linewidth]{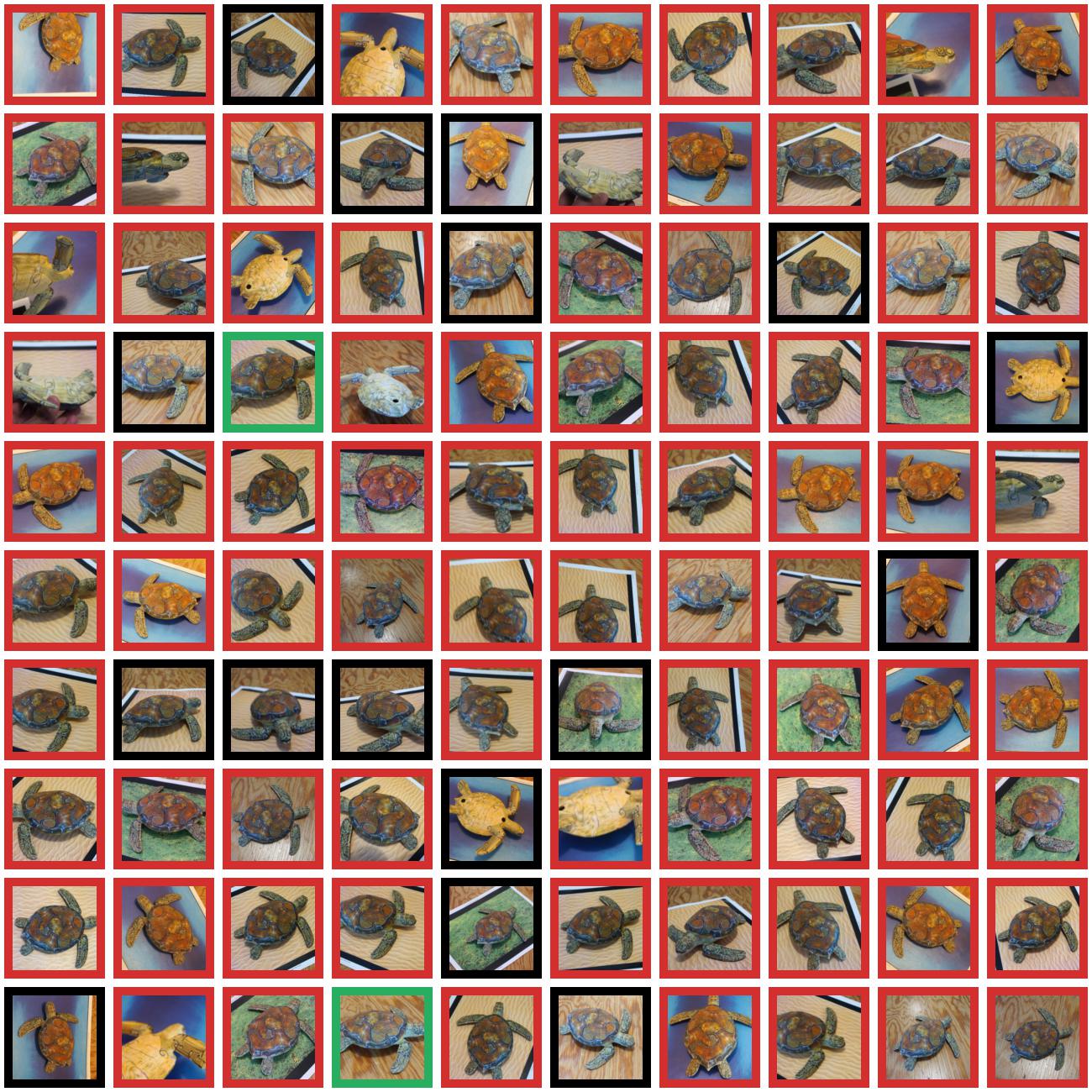}

        \makelegend{turtle}{rifle}
        \caption{
            All 100 photographs of our physical-world 3D adversarial turtle.
        }
        \label{fig:3d-turtle-full}
    \end{centering}
\end{figure}

\begin{figure}[H]
	\begin{centering}
        \includegraphics[width=\linewidth]{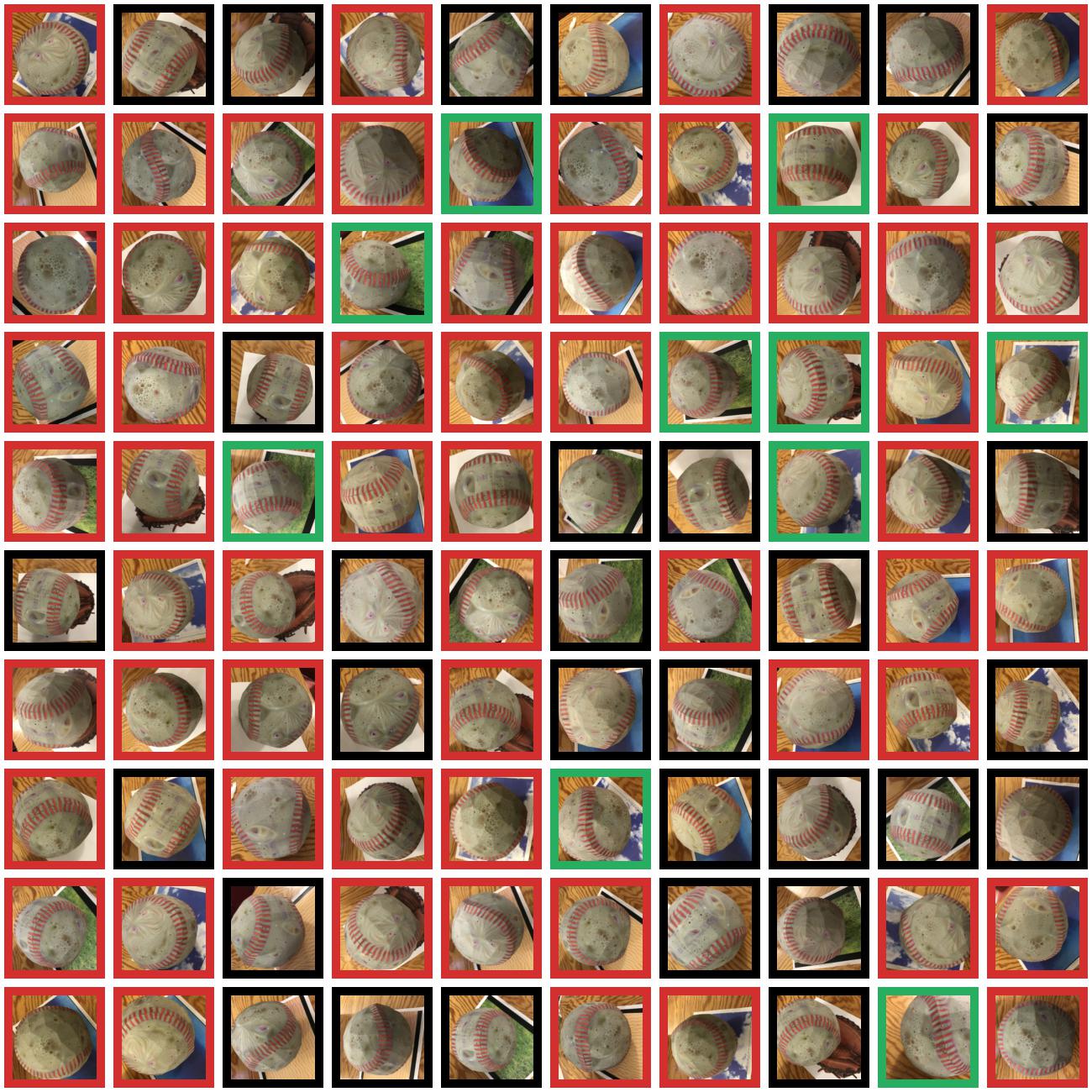}

        \makelegend{baseball}{espresso}
        \caption{
            All 100 photographs of our physical-world 3D adversarial baseball.
        }
        \label{fig:3d-baseball-full}
    \end{centering}
\end{figure}

Figure~\ref{fig:3d-turtle-full} gives all 100 photographs of our adversarial 3D-printed turtle, and Figure~\ref{fig:3d-baseball-full} gives all 100 photographs of our adversarial 3D-printed baseball.

\begin{table}[H]
	\begin{centering}
        \begin{tabular}{l r r}
            \toprule
            \textbf{Transformation} & \textbf{Minimum} & \textbf{Maximum} \\
            \midrule
            Scale & $0.9$ & $1.4$ \\
            Rotation & $-22.5^\circ$ & $22.5^\circ$ \\
            Lighten / Darken & $-0.05$ & $0.05$ \\
            Gaussian Noise (stdev) & $0.0$ & $0.1$ \\
            Translation & \multicolumn{2}{c}{any in-bounds} \\
            \bottomrule
        \end{tabular}
        \caption{
        Distribution of transformations for the 2D case, where each parameter is sampled uniformly at random from the specified range.
        }
		\label{tab:2d-dist}
    \end{centering}
\end{table}

\begin{table}[H]
	\begin{centering}
        \begin{tabular}{l r r}
            \toprule
            \textbf{Transformation} & \textbf{Minimum} & \textbf{Maximum} \\
            \midrule
            Camera distance & $2.5$ & $3.0$ \\
            X/Y translation & $-0.05$ & $0.05$ \\
            Rotation & \multicolumn{2}{c}{any} \\
            Background & (0.1, 0.1, 0.1) & (1.0, 1.0, 1.0) \\
            \bottomrule
        \end{tabular}
        \caption{
        Distribution of transformations for the 3D case when working in simulation, where each parameter is sampled uniformly at random from the specified range.
        }
		\label{tab:3d-sim-dist}
    \end{centering}
\end{table}

\begin{table}[H]
	\begin{centering}
        \begin{tabular}{l r r}
            \toprule
            \textbf{Transformation} & \textbf{Minimum} & \textbf{Maximum} \\
            \midrule
            Camera distance & $2.5$ & $3.0$ \\
            X/Y translation & $-0.05$ & $0.05$ \\
            Rotation & \multicolumn{2}{c}{any} \\
            Background & (0.1, 0.1, 0.1) & (1.0, 1.0, 1.0) \\
            Lighten / Darken (additive) & $-0.15$ & $0.15$ \\
            Lighten / Darken (multiplicative) & $0.5$ & $2.0$ \\
            Per-channel (additive) & $-0.15$ & $0.15$ \\
            Per-channel (multiplicative) & $0.7$ & $1.3$ \\
            Gaussian Noise (stdev) & $0.0$ & $0.1$ \\
            \bottomrule
        \end{tabular}
        \caption{
        Distribution of transformations for the physical-world 3D case, approximating rendering, physical-world phenomena, and printing error.
        }
		\label{tab:3d-dist}
    \end{centering}
\end{table}

\begin{figure}[H]
	\begin{centering}
        \input{./gen/2d-appendix-1.tex}
        \caption{
            A random sample of 2D adversarial examples.
        }
		\label{fig:2d-appendix-1}
    \end{centering}
\end{figure}

\begin{figure}
	\begin{centering}
        \input{./gen/2d-appendix-2.tex}
        \caption{
            A random sample of 2D adversarial examples.
        }
		\label{fig:2d-appendix-2}
    \end{centering}
\end{figure}

\begin{figure}[H]
	\begin{centering}
        \input{./gen/3d-sim-appendix-1.tex}
        \caption{
            A random sample of 3D adversarial examples.
        }
		\label{fig:3d-sim-appendix-1}
    \end{centering}
\end{figure}

\begin{figure}[H]
    \begin{centering}
        \input{./gen/3d-sim-appendix-2.tex}
        \caption{
            A random sample of 3D adversarial examples.
        }
        \label{fig:3d-sim-appendix-2}
    \end{centering}
\end{figure}

\begin{figure}[H]
    \begin{centering}
        \input{./gen/3d-sim-appendix-3.tex}
        \caption{
            A random sample of 3D adversarial examples.
        }
        \label{fig:3d-sim-appendix-3}
    \end{centering}
\end{figure}

%% file: gen/2d-appendix-1.tex
\begingroup\renewcommand*{\arraystretch}{3}
\begin{tabular}{C{0.17500\linewidth}|C{0.15625\linewidth}C{0.15625\linewidth}C{0.15625\linewidth}C{0.15625\linewidth}}
\includegraphics[align=c,width=\linewidth]{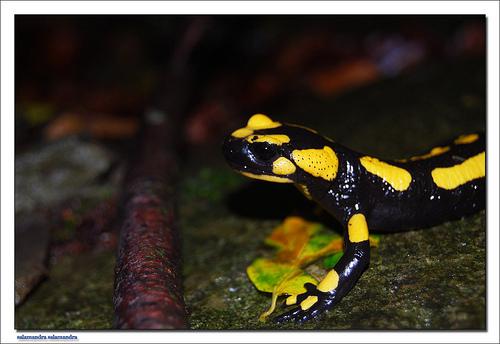} \newline Original: European fire salamander & \includegraphics[align=c,width=\linewidth]{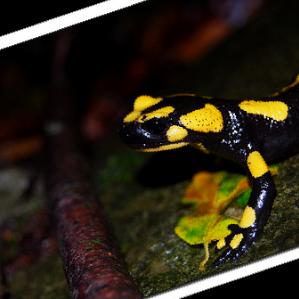} \newline $P(true)$: 93\% \newline $P(adv)$: 0\% & \includegraphics[align=c,width=\linewidth]{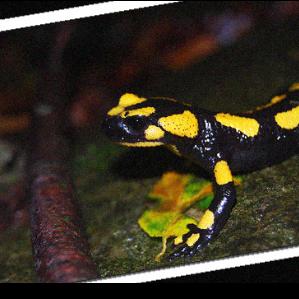} \newline $P(true)$: 91\% \newline $P(adv)$: 0\% & \includegraphics[align=c,width=\linewidth]{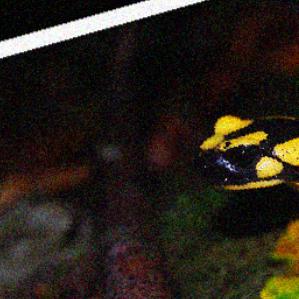} \newline $P(true)$: 93\% \newline $P(adv)$: 0\% & \includegraphics[align=c,width=\linewidth]{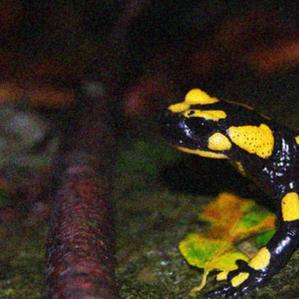} \newline $P(true)$: 93\% \newline $P(adv)$: 0\% \\ 
\includegraphics[align=c,width=\linewidth]{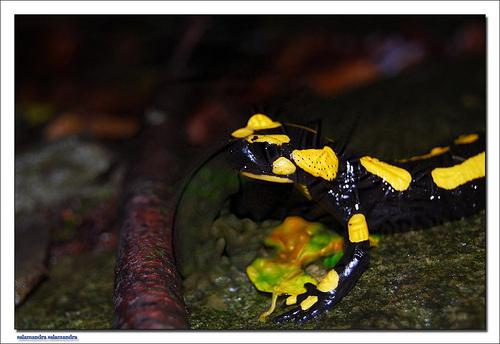} \newline Adv: guacamole& \includegraphics[align=c,width=\linewidth]{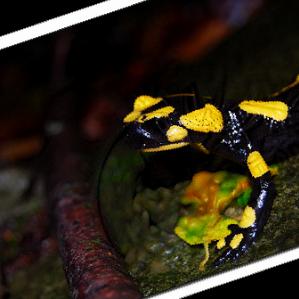} \newline $P(true)$: 0\% \newline $P(adv)$: 99\% & \includegraphics[align=c,width=\linewidth]{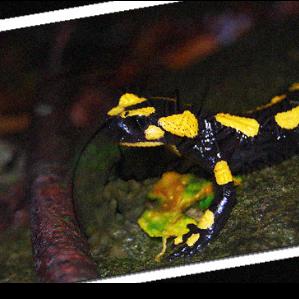} \newline $P(true)$: 0\% \newline $P(adv)$: 99\% & \includegraphics[align=c,width=\linewidth]{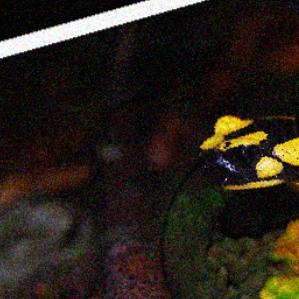} \newline $P(true)$: 0\% \newline $P(adv)$: 96\% & \includegraphics[align=c,width=\linewidth]{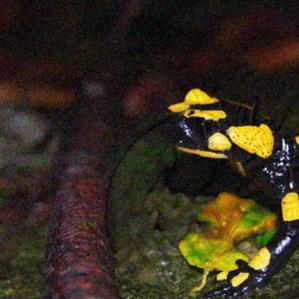} \newline $P(true)$: 0\% \newline $P(adv)$: 95\% \\ 
\includegraphics[align=c,width=\linewidth]{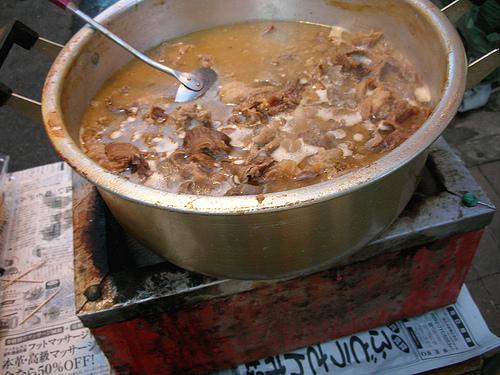} \newline Original: caldron & \includegraphics[align=c,width=\linewidth]{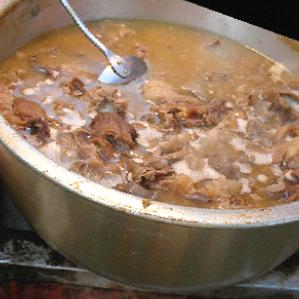} \newline $P(true)$: 75\% \newline $P(adv)$: 0\% & \includegraphics[align=c,width=\linewidth]{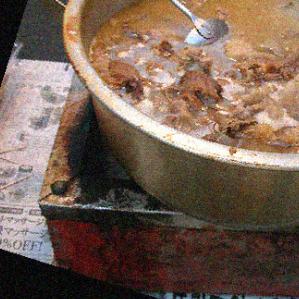} \newline $P(true)$: 83\% \newline $P(adv)$: 0\% & \includegraphics[align=c,width=\linewidth]{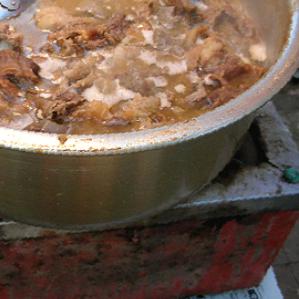} \newline $P(true)$: 54\% \newline $P(adv)$: 0\% & \includegraphics[align=c,width=\linewidth]{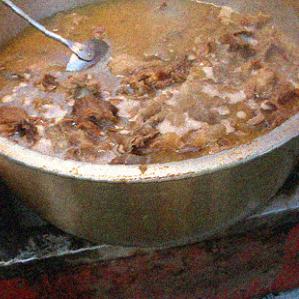} \newline $P(true)$: 80\% \newline $P(adv)$: 0\% \\ 
\includegraphics[align=c,width=\linewidth]{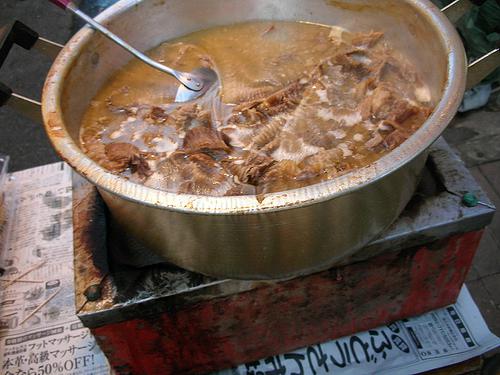} \newline Adv: velvet& \includegraphics[align=c,width=\linewidth]{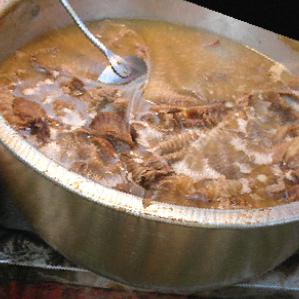} \newline $P(true)$: 0\% \newline $P(adv)$: 94\% & \includegraphics[align=c,width=\linewidth]{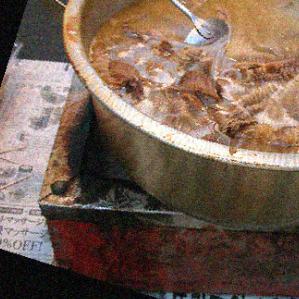} \newline $P(true)$: 0\% \newline $P(adv)$: 94\% & \includegraphics[align=c,width=\linewidth]{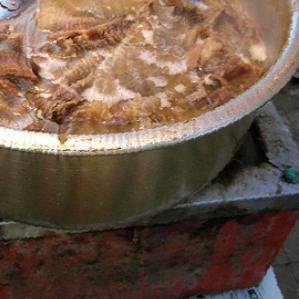} \newline $P(true)$: 1\% \newline $P(adv)$: 91\% & \includegraphics[align=c,width=\linewidth]{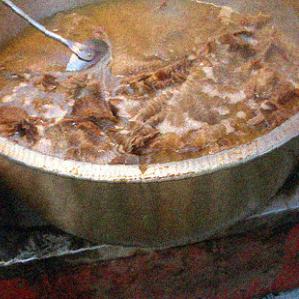} \newline $P(true)$: 0\% \newline $P(adv)$: 100\% \\ 
\includegraphics[align=c,width=\linewidth]{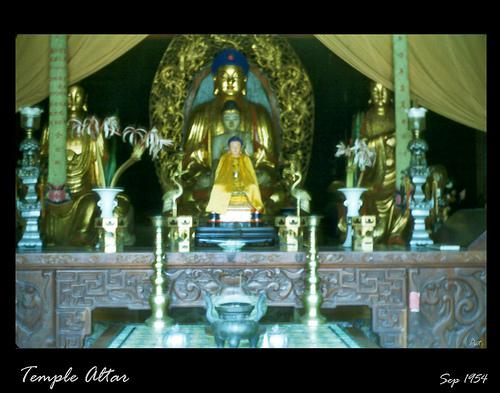} \newline Original: altar & \includegraphics[align=c,width=\linewidth]{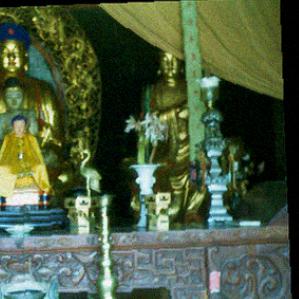} \newline $P(true)$: 87\% \newline $P(adv)$: 0\% & \includegraphics[align=c,width=\linewidth]{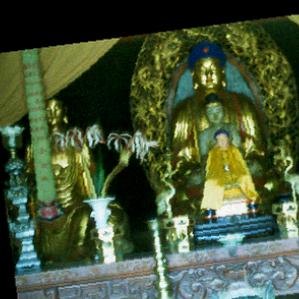} \newline $P(true)$: 38\% \newline $P(adv)$: 0\% & \includegraphics[align=c,width=\linewidth]{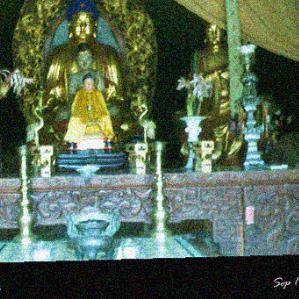} \newline $P(true)$: 59\% \newline $P(adv)$: 0\% & \includegraphics[align=c,width=\linewidth]{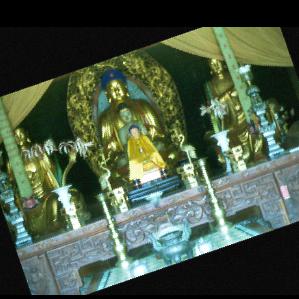} \newline $P(true)$: 2\% \newline $P(adv)$: 0\% \\ 
\includegraphics[align=c,width=\linewidth]{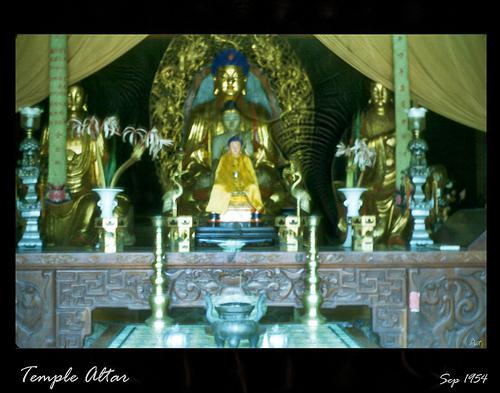} \newline Adv: African elephant& \includegraphics[align=c,width=\linewidth]{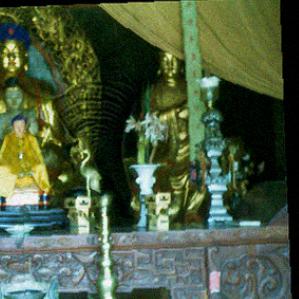} \newline $P(true)$: 0\% \newline $P(adv)$: 93\% & \includegraphics[align=c,width=\linewidth]{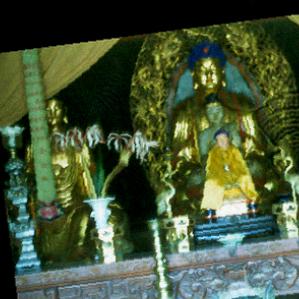} \newline $P(true)$: 0\% \newline $P(adv)$: 87\% & \includegraphics[align=c,width=\linewidth]{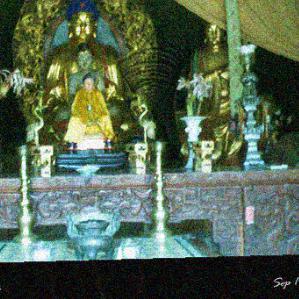} \newline $P(true)$: 3\% \newline $P(adv)$: 73\% & \includegraphics[align=c,width=\linewidth]{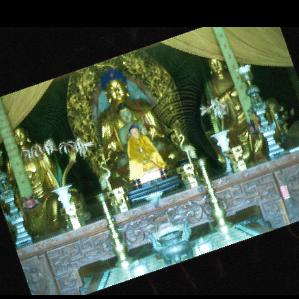} \newline $P(true)$: 0\% \newline $P(adv)$: 92\% \\ 
\end{tabular}
\endgroup

%% file: gen/2d-appendix-2.tex
\begingroup\renewcommand*{\arraystretch}{3}
\begin{tabular}{C{0.17500\linewidth}|C{0.15625\linewidth}C{0.15625\linewidth}C{0.15625\linewidth}C{0.15625\linewidth}}
\includegraphics[align=c,width=\linewidth]{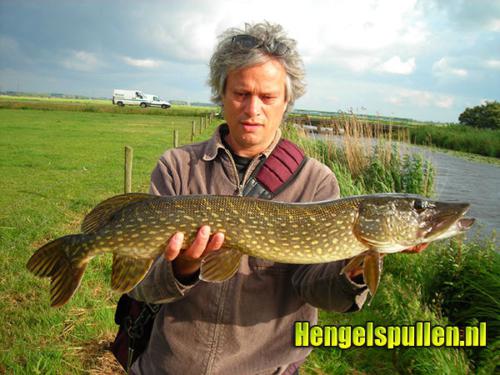} \newline Original: barracouta & \includegraphics[align=c,width=\linewidth]{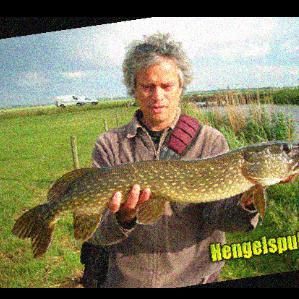} \newline $P(true)$: 91\% \newline $P(adv)$: 0\% & \includegraphics[align=c,width=\linewidth]{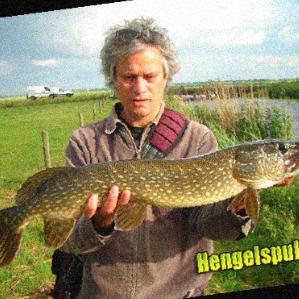} \newline $P(true)$: 95\% \newline $P(adv)$: 0\% & \includegraphics[align=c,width=\linewidth]{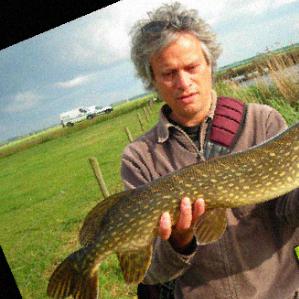} \newline $P(true)$: 92\% \newline $P(adv)$: 0\% & \includegraphics[align=c,width=\linewidth]{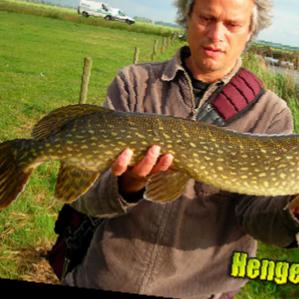} \newline $P(true)$: 92\% \newline $P(adv)$: 0\% \\ 
\includegraphics[align=c,width=\linewidth]{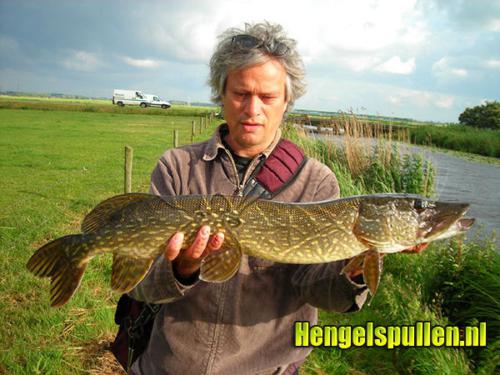} \newline Adv: tick& \includegraphics[align=c,width=\linewidth]{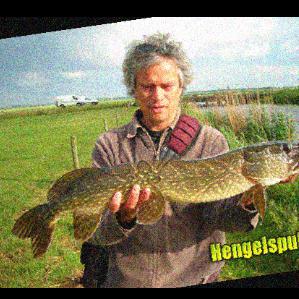} \newline $P(true)$: 0\% \newline $P(adv)$: 88\% & \includegraphics[align=c,width=\linewidth]{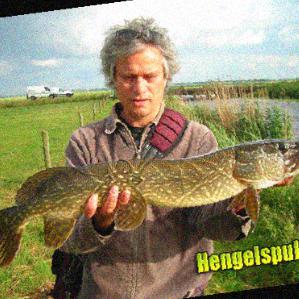} \newline $P(true)$: 0\% \newline $P(adv)$: 99\% & \includegraphics[align=c,width=\linewidth]{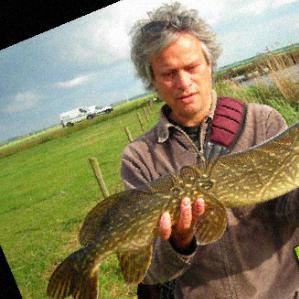} \newline $P(true)$: 0\% \newline $P(adv)$: 98\% & \includegraphics[align=c,width=\linewidth]{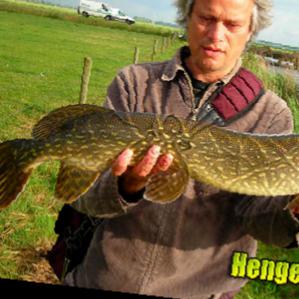} \newline $P(true)$: 0\% \newline $P(adv)$: 95\% \\ 
\includegraphics[align=c,width=\linewidth]{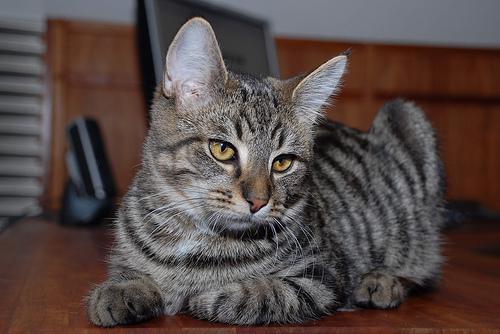} \newline Original: tiger cat & \includegraphics[align=c,width=\linewidth]{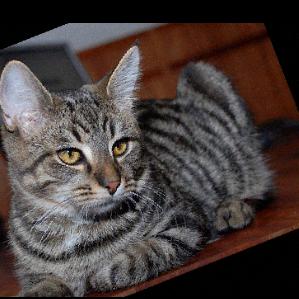} \newline $P(true)$: 85\% \newline $P(adv)$: 0\% & \includegraphics[align=c,width=\linewidth]{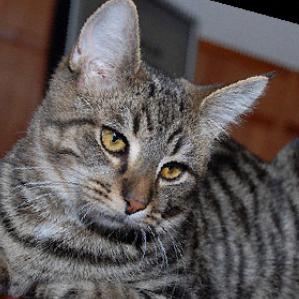} \newline $P(true)$: 91\% \newline $P(adv)$: 0\% & \includegraphics[align=c,width=\linewidth]{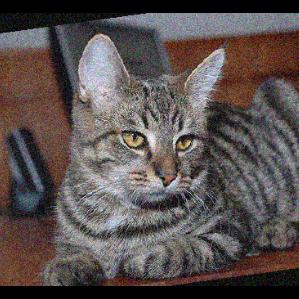} \newline $P(true)$: 69\% \newline $P(adv)$: 0\% & \includegraphics[align=c,width=\linewidth]{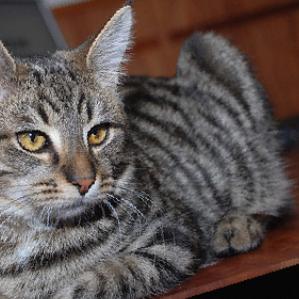} \newline $P(true)$: 96\% \newline $P(adv)$: 0\% \\ 
\includegraphics[align=c,width=\linewidth]{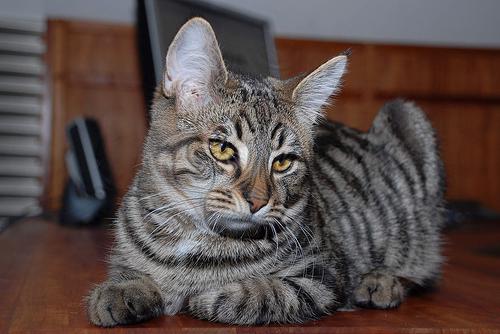} \newline Adv: tiger& \includegraphics[align=c,width=\linewidth]{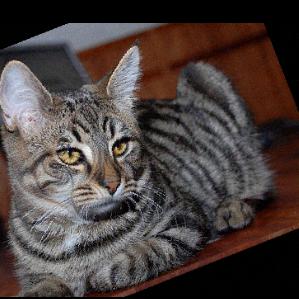} \newline $P(true)$: 32\% \newline $P(adv)$: 54\% & \includegraphics[align=c,width=\linewidth]{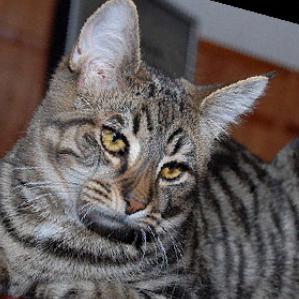} \newline $P(true)$: 11\% \newline $P(adv)$: 84\% & \includegraphics[align=c,width=\linewidth]{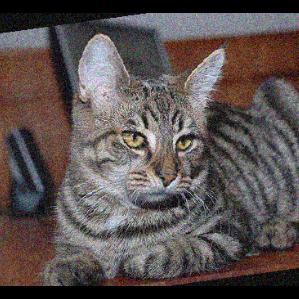} \newline $P(true)$: 59\% \newline $P(adv)$: 22\% & \includegraphics[align=c,width=\linewidth]{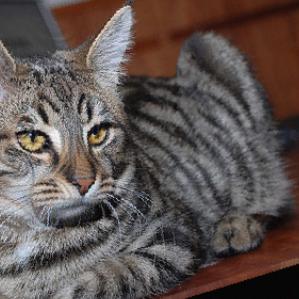} \newline $P(true)$: 14\% \newline $P(adv)$: 79\% \\ 
\includegraphics[align=c,width=\linewidth]{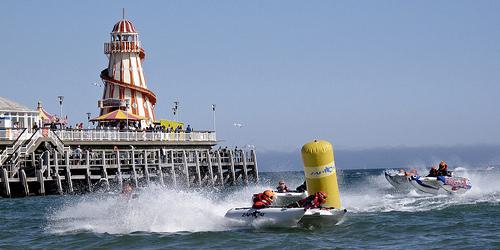} \newline Original: speedboat & \includegraphics[align=c,width=\linewidth]{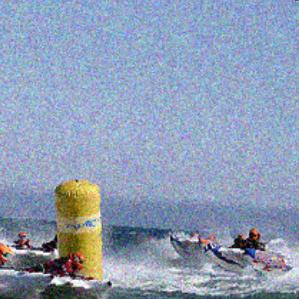} \newline $P(true)$: 14\% \newline $P(adv)$: 0\% & \includegraphics[align=c,width=\linewidth]{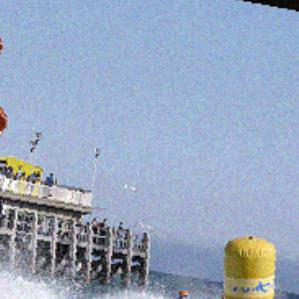} \newline $P(true)$: 1\% \newline $P(adv)$: 0\% & \includegraphics[align=c,width=\linewidth]{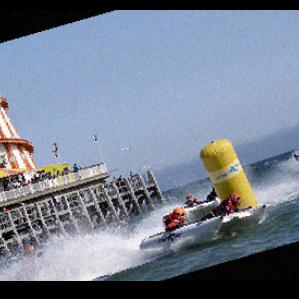} \newline $P(true)$: 1\% \newline $P(adv)$: 0\% & \includegraphics[align=c,width=\linewidth]{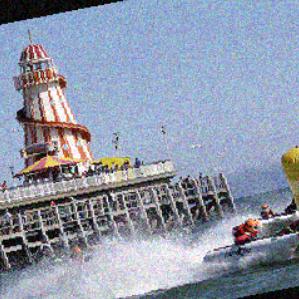} \newline $P(true)$: 1\% \newline $P(adv)$: 0\% \\ 
\includegraphics[align=c,width=\linewidth]{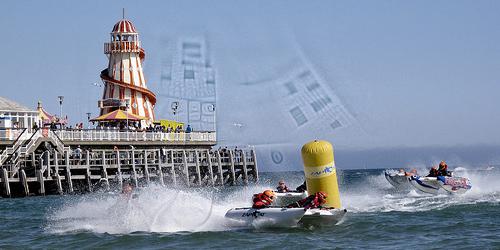} \newline Adv: crossword puzzle& \includegraphics[align=c,width=\linewidth]{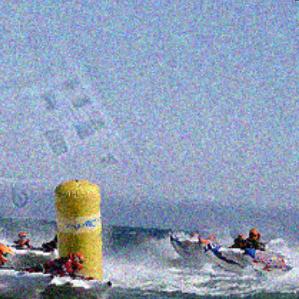} \newline $P(true)$: 3\% \newline $P(adv)$: 91\% & \includegraphics[align=c,width=\linewidth]{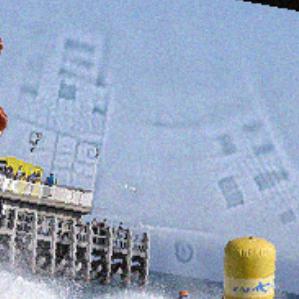} \newline $P(true)$: 0\% \newline $P(adv)$: 100\% & \includegraphics[align=c,width=\linewidth]{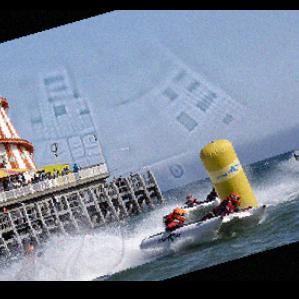} \newline $P(true)$: 0\% \newline $P(adv)$: 100\% & \includegraphics[align=c,width=\linewidth]{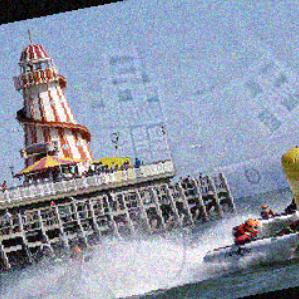} \newline $P(true)$: 0\% \newline $P(adv)$: 100\% \\ 
\end{tabular}
\endgroup

%% file: gen/3d-sim-appendix-1.tex
\begingroup\renewcommand*{\arraystretch}{3}
\begin{tabular}{C{0.17500\linewidth}|C{0.15625\linewidth}C{0.15625\linewidth}C{0.15625\linewidth}C{0.15625\linewidth}}
Original: barrel \vspace{8ex}& \includegraphics[align=c,width=\linewidth]{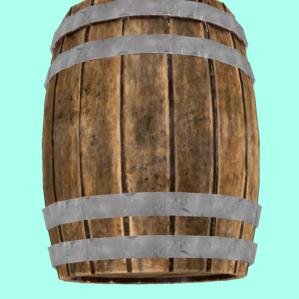} \newline $P(true)$: 96\% \newline $P(adv)$: 0\% & \includegraphics[align=c,width=\linewidth]{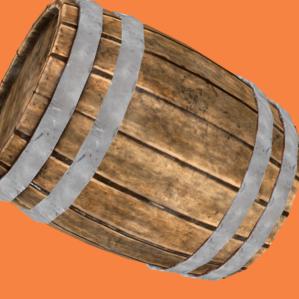} \newline $P(true)$: 99\% \newline $P(adv)$: 0\% & \includegraphics[align=c,width=\linewidth]{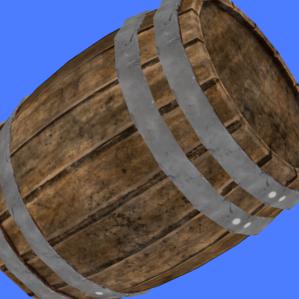} \newline $P(true)$: 96\% \newline $P(adv)$: 0\% & \includegraphics[align=c,width=\linewidth]{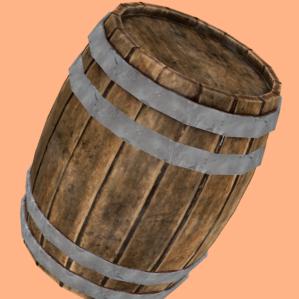} \newline $P(true)$: 97\% \newline $P(adv)$: 0\% \\ 
Adv: guillotine& \includegraphics[align=c,width=\linewidth]{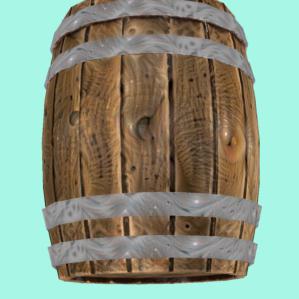} \newline $P(true)$: 1\% \newline $P(adv)$: 10\% & \includegraphics[align=c,width=\linewidth]{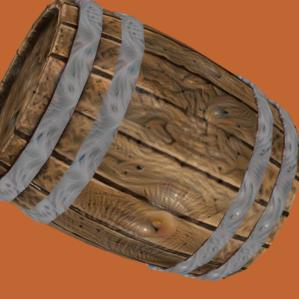} \newline $P(true)$: 0\% \newline $P(adv)$: 95\% & \includegraphics[align=c,width=\linewidth]{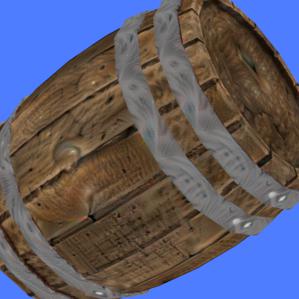} \newline $P(true)$: 0\% \newline $P(adv)$: 91\% & \includegraphics[align=c,width=\linewidth]{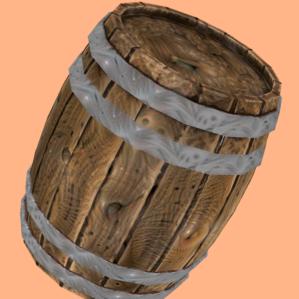} \newline $P(true)$: 3\% \newline $P(adv)$: 4\% \\ 
Original: baseball \vspace{8ex}& \includegraphics[align=c,width=\linewidth]{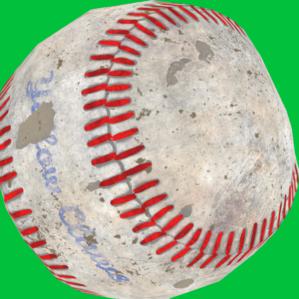} \newline $P(true)$: 100\% \newline $P(adv)$: 0\% & \includegraphics[align=c,width=\linewidth]{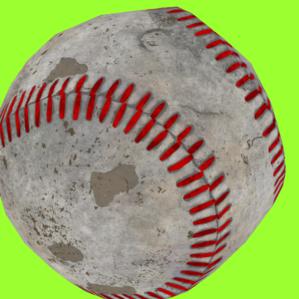} \newline $P(true)$: 100\% \newline $P(adv)$: 0\% & \includegraphics[align=c,width=\linewidth]{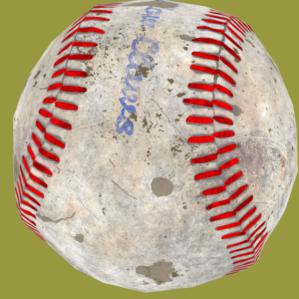} \newline $P(true)$: 100\% \newline $P(adv)$: 0\% & \includegraphics[align=c,width=\linewidth]{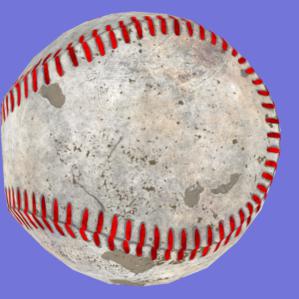} \newline $P(true)$: 100\% \newline $P(adv)$: 0\% \\ 
Adv: green lizard& \includegraphics[align=c,width=\linewidth]{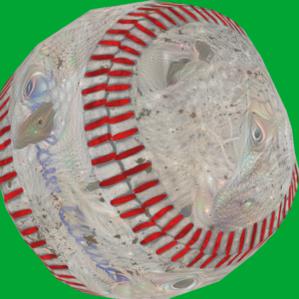} \newline $P(true)$: 0\% \newline $P(adv)$: 66\% & \includegraphics[align=c,width=\linewidth]{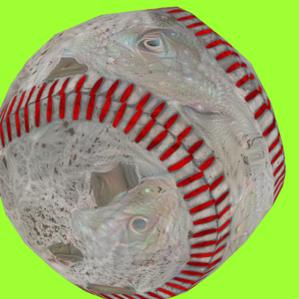} \newline $P(true)$: 0\% \newline $P(adv)$: 94\% & \includegraphics[align=c,width=\linewidth]{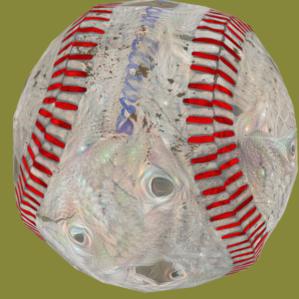} \newline $P(true)$: 0\% \newline $P(adv)$: 87\% & \includegraphics[align=c,width=\linewidth]{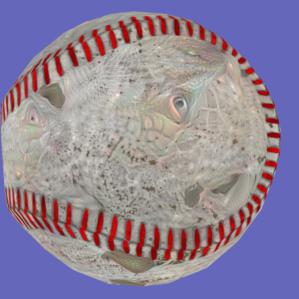} \newline $P(true)$: 0\% \newline $P(adv)$: 94\% \\ 
Original: turtle \vspace{8ex}& \includegraphics[align=c,width=\linewidth]{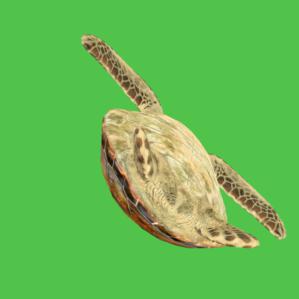} \newline $P(true)$: 94\% \newline $P(adv)$: 0\% & \includegraphics[align=c,width=\linewidth]{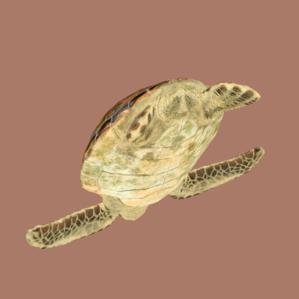} \newline $P(true)$: 98\% \newline $P(adv)$: 0\% & \includegraphics[align=c,width=\linewidth]{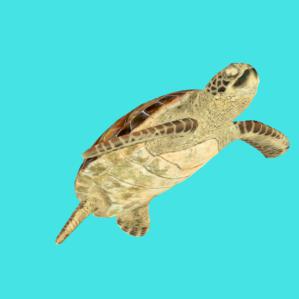} \newline $P(true)$: 90\% \newline $P(adv)$: 0\% & \includegraphics[align=c,width=\linewidth]{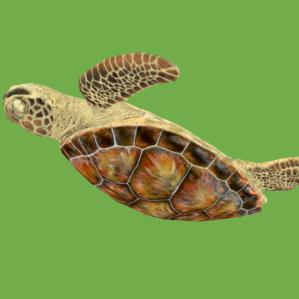} \newline $P(true)$: 97\% \newline $P(adv)$: 0\% \\ 
Adv: Bouvier des Flandres& \includegraphics[align=c,width=\linewidth]{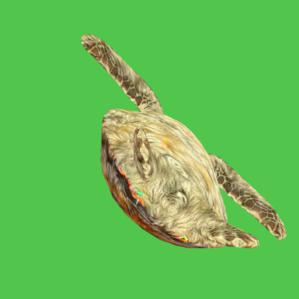} \newline $P(true)$: 1\% \newline $P(adv)$: 1\% & \includegraphics[align=c,width=\linewidth]{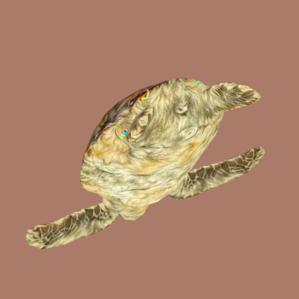} \newline $P(true)$: 0\% \newline $P(adv)$: 6\% & \includegraphics[align=c,width=\linewidth]{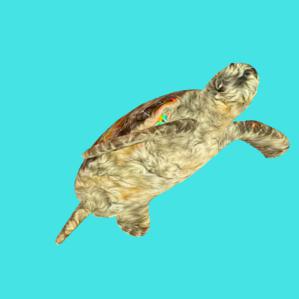} \newline $P(true)$: 0\% \newline $P(adv)$: 21\% & \includegraphics[align=c,width=\linewidth]{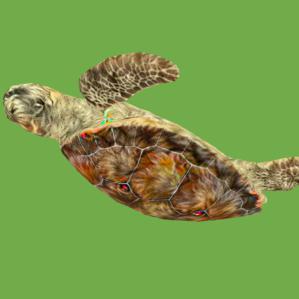} \newline $P(true)$: 0\% \newline $P(adv)$: 84\% \\ 
\end{tabular}
\endgroup

%% file: gen/3d-sim-appendix-2.tex
\begingroup\renewcommand*{\arraystretch}{3}
\begin{tabular}{C{0.17500\linewidth}|C{0.15625\linewidth}C{0.15625\linewidth}C{0.15625\linewidth}C{0.15625\linewidth}}
Original: baseball \vspace{8ex}& \includegraphics[align=c,width=\linewidth]{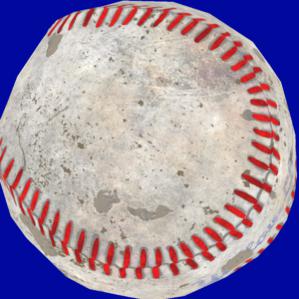} \newline $P(true)$: 100\% \newline $P(adv)$: 0\% & \includegraphics[align=c,width=\linewidth]{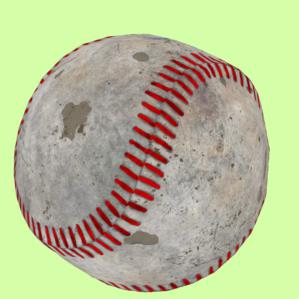} \newline $P(true)$: 100\% \newline $P(adv)$: 0\% & \includegraphics[align=c,width=\linewidth]{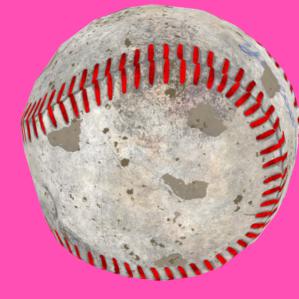} \newline $P(true)$: 100\% \newline $P(adv)$: 0\% & \includegraphics[align=c,width=\linewidth]{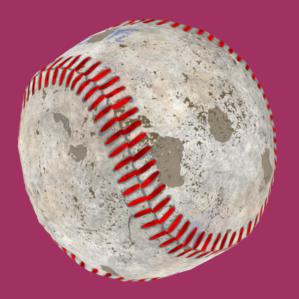} \newline $P(true)$: 100\% \newline $P(adv)$: 0\% \\ 
Adv: Airedale& \includegraphics[align=c,width=\linewidth]{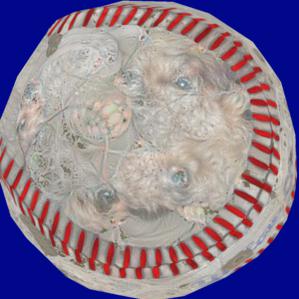} \newline $P(true)$: 0\% \newline $P(adv)$: 94\% & \includegraphics[align=c,width=\linewidth]{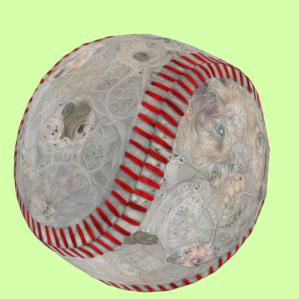} \newline $P(true)$: 0\% \newline $P(adv)$: 6\% & \includegraphics[align=c,width=\linewidth]{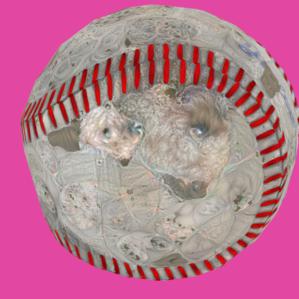} \newline $P(true)$: 0\% \newline $P(adv)$: 96\% & \includegraphics[align=c,width=\linewidth]{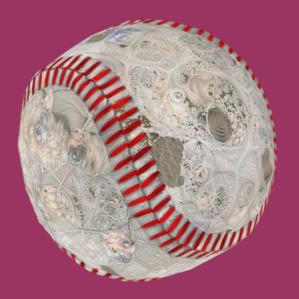} \newline $P(true)$: 0\% \newline $P(adv)$: 18\% \\ 
Original: orange \vspace{8ex}& \includegraphics[align=c,width=\linewidth]{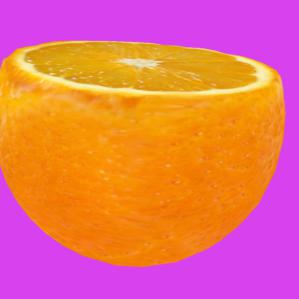} \newline $P(true)$: 73\% \newline $P(adv)$: 0\% & \includegraphics[align=c,width=\linewidth]{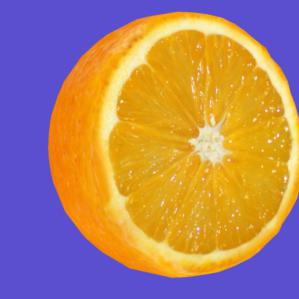} \newline $P(true)$: 29\% \newline $P(adv)$: 0\% & \includegraphics[align=c,width=\linewidth]{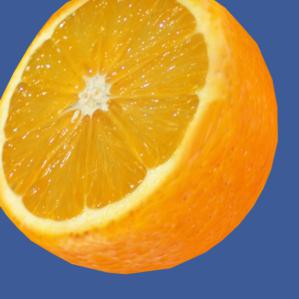} \newline $P(true)$: 20\% \newline $P(adv)$: 0\% & \includegraphics[align=c,width=\linewidth]{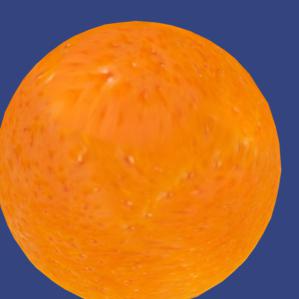} \newline $P(true)$: 85\% \newline $P(adv)$: 0\% \\ 
Adv: power drill& \includegraphics[align=c,width=\linewidth]{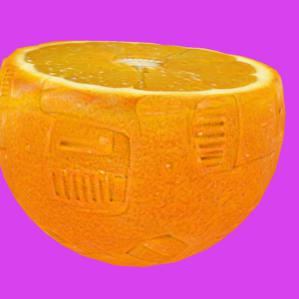} \newline $P(true)$: 0\% \newline $P(adv)$: 89\% & \includegraphics[align=c,width=\linewidth]{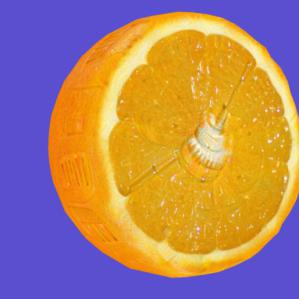} \newline $P(true)$: 4\% \newline $P(adv)$: 75\% & \includegraphics[align=c,width=\linewidth]{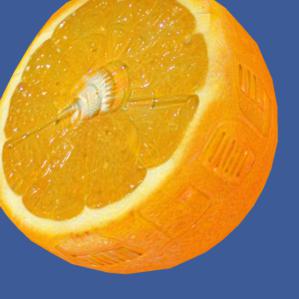} \newline $P(true)$: 0\% \newline $P(adv)$: 98\% & \includegraphics[align=c,width=\linewidth]{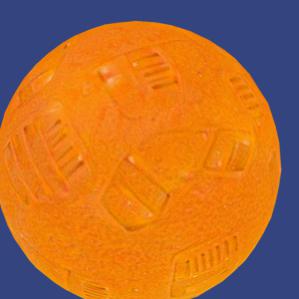} \newline $P(true)$: 0\% \newline $P(adv)$: 84\% \\ 
Original: dog \vspace{8ex}& \includegraphics[align=c,width=\linewidth]{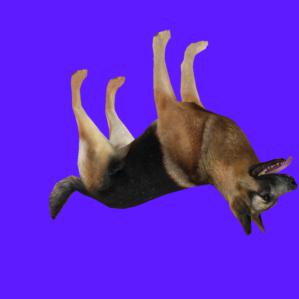} \newline $P(true)$: 1\% \newline $P(adv)$: 0\% & \includegraphics[align=c,width=\linewidth]{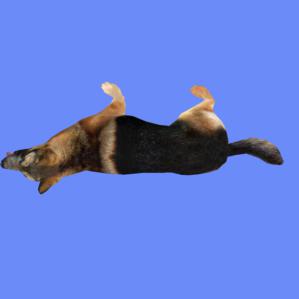} \newline $P(true)$: 32\% \newline $P(adv)$: 0\% & \includegraphics[align=c,width=\linewidth]{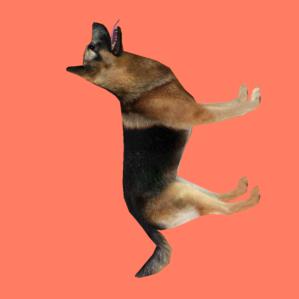} \newline $P(true)$: 12\% \newline $P(adv)$: 0\% & \includegraphics[align=c,width=\linewidth]{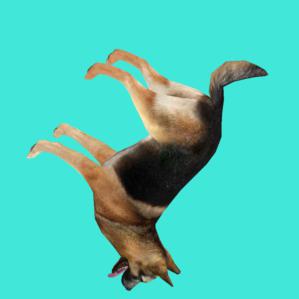} \newline $P(true)$: 0\% \newline $P(adv)$: 0\% \\ 
Adv: bittern& \includegraphics[align=c,width=\linewidth]{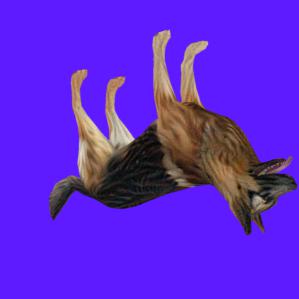} \newline $P(true)$: 0\% \newline $P(adv)$: 97\% & \includegraphics[align=c,width=\linewidth]{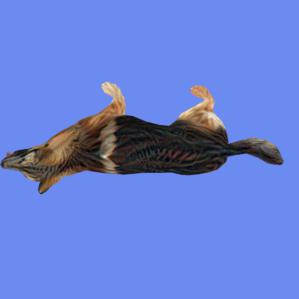} \newline $P(true)$: 0\% \newline $P(adv)$: 91\% & \includegraphics[align=c,width=\linewidth]{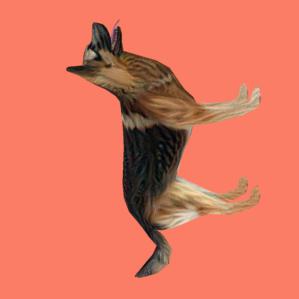} \newline $P(true)$: 0\% \newline $P(adv)$: 98\% & \includegraphics[align=c,width=\linewidth]{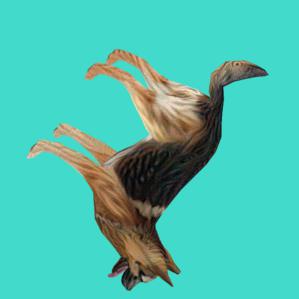} \newline $P(true)$: 0\% \newline $P(adv)$: 97\% \\ 
\end{tabular}
\endgroup

%% file: gen/3d-sim-appendix-3.tex
\begingroup\renewcommand*{\arraystretch}{3}
\begin{tabular}{C{0.17500\linewidth}|C{0.15625\linewidth}C{0.15625\linewidth}C{0.15625\linewidth}C{0.15625\linewidth}}
Original: teddybear \vspace{8ex}& \includegraphics[align=c,width=\linewidth]{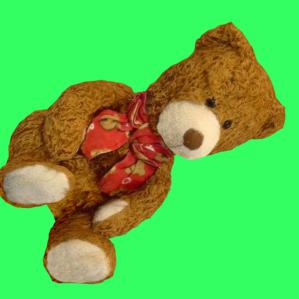} \newline $P(true)$: 90\% \newline $P(adv)$: 0\% & \includegraphics[align=c,width=\linewidth]{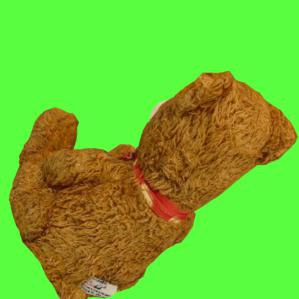} \newline $P(true)$: 1\% \newline $P(adv)$: 0\% & \includegraphics[align=c,width=\linewidth]{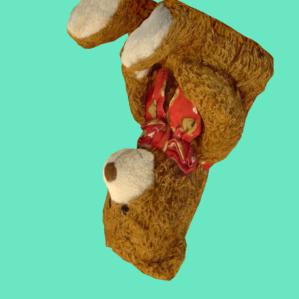} \newline $P(true)$: 98\% \newline $P(adv)$: 0\% & \includegraphics[align=c,width=\linewidth]{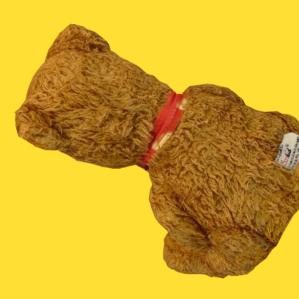} \newline $P(true)$: 5\% \newline $P(adv)$: 0\% \\ 
Adv: sock& \includegraphics[align=c,width=\linewidth]{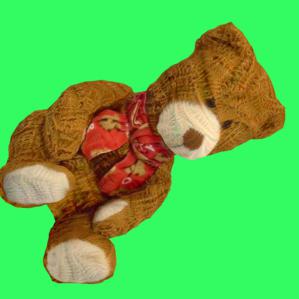} \newline $P(true)$: 0\% \newline $P(adv)$: 99\% & \includegraphics[align=c,width=\linewidth]{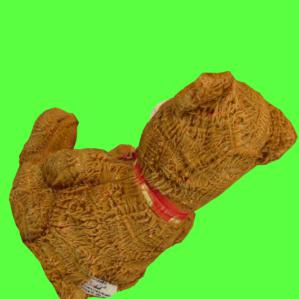} \newline $P(true)$: 0\% \newline $P(adv)$: 99\% & \includegraphics[align=c,width=\linewidth]{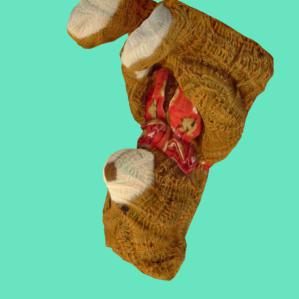} \newline $P(true)$: 0\% \newline $P(adv)$: 98\% & \includegraphics[align=c,width=\linewidth]{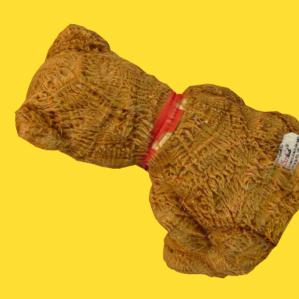} \newline $P(true)$: 0\% \newline $P(adv)$: 99\% \\ 
Original: clownfish \vspace{8ex}& \includegraphics[align=c,width=\linewidth]{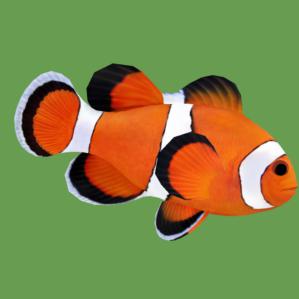} \newline $P(true)$: 46\% \newline $P(adv)$: 0\% & \includegraphics[align=c,width=\linewidth]{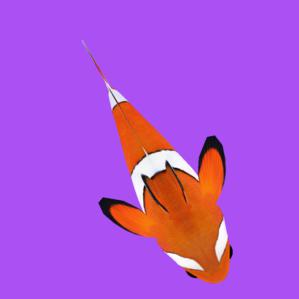} \newline $P(true)$: 14\% \newline $P(adv)$: 0\% & \includegraphics[align=c,width=\linewidth]{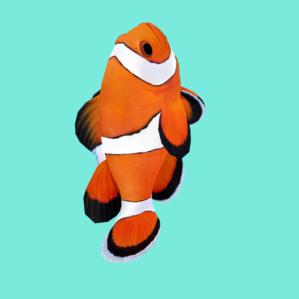} \newline $P(true)$: 2\% \newline $P(adv)$: 0\% & \includegraphics[align=c,width=\linewidth]{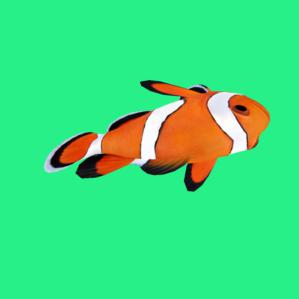} \newline $P(true)$: 65\% \newline $P(adv)$: 0\% \\ 
Adv: panpipe& \includegraphics[align=c,width=\linewidth]{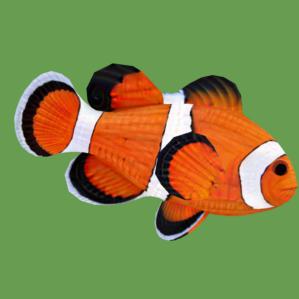} \newline $P(true)$: 0\% \newline $P(adv)$: 100\% & \includegraphics[align=c,width=\linewidth]{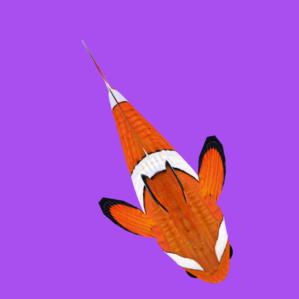} \newline $P(true)$: 0\% \newline $P(adv)$: 1\% & \includegraphics[align=c,width=\linewidth]{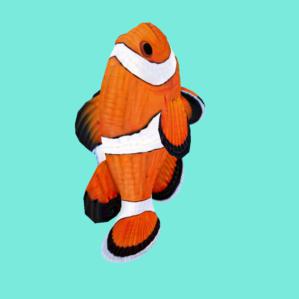} \newline $P(true)$: 0\% \newline $P(adv)$: 12\% & \includegraphics[align=c,width=\linewidth]{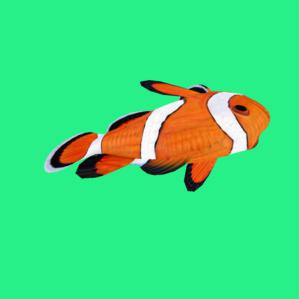} \newline $P(true)$: 0\% \newline $P(adv)$: 0\% \\ 
Original: sofa \vspace{8ex}& \includegraphics[align=c,width=\linewidth]{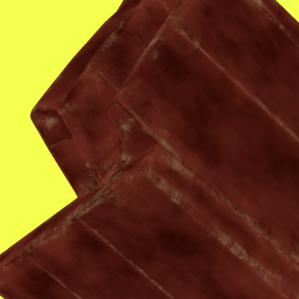} \newline $P(true)$: 15\% \newline $P(adv)$: 0\% & \includegraphics[align=c,width=\linewidth]{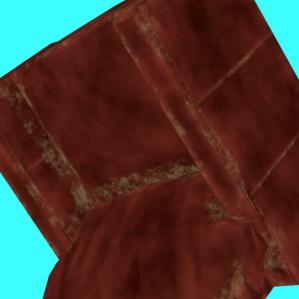} \newline $P(true)$: 73\% \newline $P(adv)$: 0\% & \includegraphics[align=c,width=\linewidth]{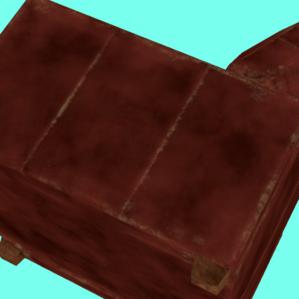} \newline $P(true)$: 1\% \newline $P(adv)$: 0\% & \includegraphics[align=c,width=\linewidth]{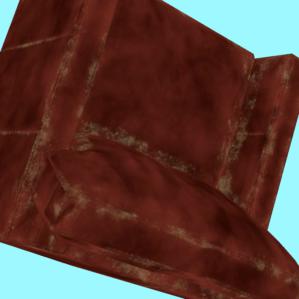} \newline $P(true)$: 70\% \newline $P(adv)$: 0\% \\ 
Adv: sturgeon& \includegraphics[align=c,width=\linewidth]{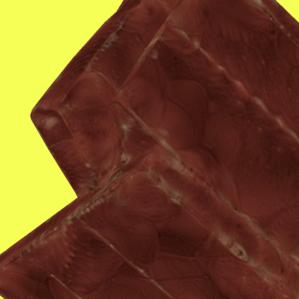} \newline $P(true)$: 0\% \newline $P(adv)$: 100\% & \includegraphics[align=c,width=\linewidth]{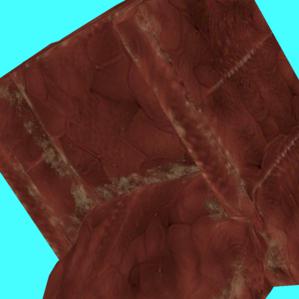} \newline $P(true)$: 0\% \newline $P(adv)$: 100\% & \includegraphics[align=c,width=\linewidth]{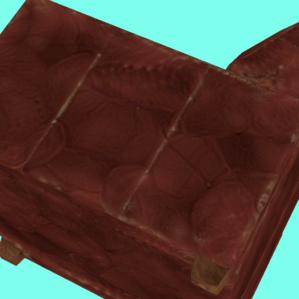} \newline $P(true)$: 0\% \newline $P(adv)$: 100\% & \includegraphics[align=c,width=\linewidth]{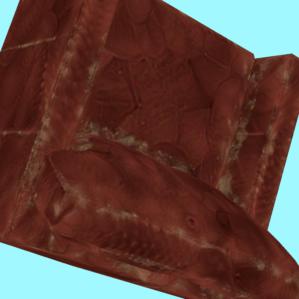} \newline $P(true)$: 0\% \newline $P(adv)$: 100\% \\ 
\end{tabular}
\endgroup

%% file: arxiv.bbl
\begin{thebibliography}{36}
\providecommand{\natexlab}[1]{#1}
\providecommand{\url}[1]{\texttt{#1}}
\expandafter\ifx\csname urlstyle\endcsname\relax
  \providecommand{\doi}[1]{doi: #1}\else
  \providecommand{\doi}{doi: \begingroup \urlstyle{rm}\Url}\fi

\bibitem[Athalye et~al.(2018)Athalye, Carlini, and Wagner]{anish-carlini}
Athalye, A., Carlini, N., and Wagner, D.
\newblock Obfuscated gradients give a false sense of security: Circumventing
  defenses to adversarial examples.
\newblock 2018.
\newblock URL \url{https://arxiv.org/abs/1802.00420}.

\bibitem[Biggio et~al.(2013)Biggio, Corona, Maiorca, Nelson, {\v{S}}rndi{\'c},
  Laskov, Giacinto, and Roli]{biggio2013evasion}
Biggio, B., Corona, I., Maiorca, D., Nelson, B., {\v{S}}rndi{\'c}, N., Laskov,
  P., Giacinto, G., and Roli, F.
\newblock Evasion attacks against machine learning at test time.
\newblock In \emph{Joint European Conference on Machine Learning and Knowledge
  Discovery in Databases}, pp.\  387--402. Springer, 2013.

\bibitem[Brown et~al.(2016)Brown, Man{\'e}, Roy, Abadi, and
  Gilmer]{brown2017patch}
Brown, T.~B., Man{\'e}, D., Roy, A., Abadi, M., and Gilmer, J.
\newblock Defensive distillation is not robust to adversarial examples.
\newblock 2016.
\newblock URL \url{https://arxiv.org/abs/1607.04311}.

\bibitem[Buckman et~al.(2018)Buckman, Roy, Raffel, and
  Goodfellow]{buckman2018thermometer}
Buckman, J., Roy, A., Raffel, C., and Goodfellow, I.
\newblock Thermometer encoding: One hot way to resist adversarial examples.
\newblock \emph{International Conference on Learning Representations}, 2018.
\newblock URL \url{https://openreview.net/forum?id=S18Su--CW}.
\newblock accepted as poster.

\bibitem[Carlini \& Wagner(2016)Carlini and Wagner]{carlini2016distillation}
Carlini, N. and Wagner, D.
\newblock Defensive distillation is not robust to adversarial examples.
\newblock 2016.
\newblock URL \url{https://arxiv.org/abs/1607.04311}.

\bibitem[Carlini \& Wagner(2017{\natexlab{a}})Carlini and
  Wagner]{carlini2017adversarial}
Carlini, N. and Wagner, D.
\newblock Adversarial examples are not easily detected: Bypassing ten detection
  methods.
\newblock \emph{AISec}, 2017{\natexlab{a}}.

\bibitem[Carlini \& Wagner(2017{\natexlab{b}})Carlini and
  Wagner]{carlini2017magnet}
Carlini, N. and Wagner, D.
\newblock Magnet and ``efficient defenses against adversarial attacks'' are not
  robust to adversarial examples.
\newblock \emph{arXiv preprint arXiv:1711.08478}, 2017{\natexlab{b}}.

\bibitem[Carlini \& Wagner(2017{\natexlab{c}})Carlini and
  Wagner]{sp2017:carlini}
Carlini, N. and Wagner, D.
\newblock Towards evaluating the robustness of neural networks.
\newblock In \emph{IEEE Symposium on Security \& Privacy}, 2017{\natexlab{c}}.

\bibitem[Carlini et~al.(2016)Carlini, Mishra, Vaidya, Zhang, Sherr, Shields,
  Wagner, and Zhou]{carlini2016hidden}
Carlini, N., Mishra, P., Vaidya, T., Zhang, Y., Sherr, M., Shields, C., Wagner,
  D., and Zhou, W.
\newblock Hidden voice commands.
\newblock In \emph{25th {USENIX} Security Symposium ({USENIX} Security 16)},
  pp.\  513--530, Austin, TX, 2016. {USENIX} Association.
\newblock ISBN 978-1-931971-32-4.
\newblock URL
  \url{https://www.usenix.org/conference/usenixsecurity16/technical-sessions/presentation/carlini}.

\bibitem[Chen et~al.(2017)Chen, Zhang, Sharma, Yi, and Hsieh]{zoo}
Chen, P.-Y., Zhang, H., Sharma, Y., Yi, J., and Hsieh, C.-J.
\newblock Zoo: Zeroth order optimization based black-box attacks to deep neural
  networks without training substitute models.
\newblock In \emph{Proceedings of the 10th ACM Workshop on Artificial
  Intelligence and Security}, AISec '17, pp.\  15--26, New York, NY, USA, 2017.
  ACM.
\newblock ISBN 978-1-4503-5202-4.
\newblock \doi{10.1145/3128572.3140448}.
\newblock URL \url{http://doi.acm.org/10.1145/3128572.3140448}.

\bibitem[Dhillon et~al.(2018)Dhillon, Azizzadenesheli, Bernstein, Kossaifi,
  Khanna, Lipton, and Anandkumar]{dhillon2018stochastic}
Dhillon, G.~S., Azizzadenesheli, K., Bernstein, J.~D., Kossaifi, J., Khanna,
  A., Lipton, Z.~C., and Anandkumar, A.
\newblock Stochastic activation pruning for robust adversarial defense.
\newblock \emph{International Conference on Learning Representations}, 2018.
\newblock URL \url{https://openreview.net/forum?id=H1uR4GZRZ}.
\newblock accepted as poster.

\bibitem[Evtimov et~al.(2017)Evtimov, Eykholt, Fernandes, Kohno, Li, Prakash,
  Rahmati, and Song]{evtimov-roadsigns}
Evtimov, I., Eykholt, K., Fernandes, E., Kohno, T., Li, B., Prakash, A.,
  Rahmati, A., and Song, D.
\newblock {Robust Physical-World Attacks on Deep Learning Models}.
\newblock 2017.
\newblock URL \url{https://arxiv.org/abs/1707.08945}.

\bibitem[Goodfellow et~al.(2015)Goodfellow, Shlens, and
  Szegedy]{iclr2015:goodfellow}
Goodfellow, I.~J., Shlens, J., and Szegedy, C.
\newblock Explaining and harnessing adversarial examples.
\newblock In \emph{Proceedings of the International Conference on Learning
  Representations (ICLR)}, 2015.

\bibitem[Guo et~al.(2018)Guo, Rana, Cisse, and van~der
  Maaten]{guo2018countering}
Guo, C., Rana, M., Cisse, M., and van~der Maaten, L.
\newblock Countering adversarial images using input transformations.
\newblock \emph{International Conference on Learning Representations}, 2018.
\newblock URL \url{https://openreview.net/forum?id=SyJ7ClWCb}.
\newblock accepted as poster.

\bibitem[Hendrik~Metzen et~al.(2017)Hendrik~Metzen, Genewein, Fischer, and
  Bischoff]{hendrik2017detecting}
Hendrik~Metzen, J., Genewein, T., Fischer, V., and Bischoff, B.
\newblock On detecting adversarial perturbations.
\newblock In \emph{International Conference on Learning Representations}, 2017.

\bibitem[Hendrycks \& Gimpel(2017)Hendrycks and Gimpel]{hendrycks2017early}
Hendrycks, D. and Gimpel, K.
\newblock Early methods for detecting adversarial images.
\newblock In \emph{International Conference on Learning Representations
  (Workshop Track)}, 2017.

\bibitem[Kurakin et~al.(2016)Kurakin, Goodfellow, and
  Bengio]{goodfellow-physical}
Kurakin, A., Goodfellow, I., and Bengio, S.
\newblock Adversarial examples in the physical world.
\newblock 2016.
\newblock URL \url{https://arxiv.org/abs/1607.02533}.

\bibitem[Lu et~al.(2017)Lu, Sibai, Fabry, and Forsyth]{lu-noneed}
Lu, J., Sibai, H., Fabry, E., and Forsyth, D.
\newblock No need to worry about adversarial examples in object detection in
  autonomous vehicles.
\newblock 2017.
\newblock URL \url{https://arxiv.org/abs/1707.03501}.

\bibitem[Luo et~al.(2016)Luo, Boix, Roig, Poggio, and Zhao]{luo-foveation}
Luo, Y., Boix, X., Roig, G., Poggio, T., and Zhao, Q.
\newblock Foveation-based mechanisms alleviate adversarial examples.
\newblock 2016.
\newblock URL \url{https://arxiv.org/abs/1511.06292}.

\bibitem[Ma et~al.(2018)Ma, Li, Wang, Erfani, Wijewickrema, Schoenebeck, Houle,
  Song, and Bailey]{ma2018characterizing}
Ma, X., Li, B., Wang, Y., Erfani, S.~M., Wijewickrema, S., Schoenebeck, G.,
  Houle, M.~E., Song, D., and Bailey, J.
\newblock Characterizing adversarial subspaces using local intrinsic
  dimensionality.
\newblock \emph{International Conference on Learning Representations}, 2018.
\newblock URL \url{https://openreview.net/forum?id=B1gJ1L2aW}.
\newblock accepted as oral presentation.

\bibitem[Madry et~al.(2017)Madry, Makelov, Schmidt, Tsipras, and
  Vladu]{madry-adversarial}
Madry, A., Makelov, A., Schmidt, L., Tsipras, D., and Vladu, A.
\newblock Towards deep learning models resistant to adversarial attacks.
\newblock 2017.
\newblock URL \url{https://arxiv.org/abs/1706.06083}.

\bibitem[McLaren(1976)]{mclaren1976cielab}
McLaren, K.
\newblock Xiii—the development of the cie 1976 (l* a* b*) uniform colour
  space and colour‐difference formula.
\newblock \emph{Journal of the Society of Dyers and Colourists}, 92\penalty0
  (9):\penalty0 338--341, September 1976.
\newblock \doi{10.1111/j.1478-4408.1976.tb03301.x}.
\newblock URL
  \url{https://onlinelibrary.wiley.com/doi/abs/10.1111/j.1478-4408.1976.tb03301.x}.

\bibitem[Meng \& Chen(2017)Meng and Chen]{meng2017magnet}
Meng, D. and Chen, H.
\newblock {MagNet}: a two-pronged defense against adversarial examples.
\newblock In \emph{ACM Conference on Computer and Communications Security
  (CCS)}, 2017.
\newblock arXiv preprint arXiv:1705.09064.

\bibitem[Moosavi{-}Dezfooli et~al.(2015)Moosavi{-}Dezfooli, Fawzi, and
  Frossard]{Moosavi-Dezfooli15}
Moosavi{-}Dezfooli, S., Fawzi, A., and Frossard, P.
\newblock Deepfool: a simple and accurate method to fool deep neural networks.
\newblock \emph{CoRR}, abs/1511.04599, 2015.
\newblock URL \url{http://arxiv.org/abs/1511.04599}.

\bibitem[Moosavi-Dezfooli et~al.(2017)Moosavi-Dezfooli, Fawzi, Fawzi, and
  Frossard]{cvpr2017:dezfooli}
Moosavi-Dezfooli, S.-M., Fawzi, A., Fawzi, O., and Frossard, P.
\newblock Universal adversarial perturbations.
\newblock In \emph{IEEE Conference on Computer Vision and Pattern Recognition
  (CVPR)}, 2017.

\bibitem[Papernot et~al.(2016{\natexlab{a}})Papernot, McDaniel, and
  Goodfellow]{papernot-transferability}
Papernot, N., McDaniel, P., and Goodfellow, I.
\newblock Transferability in machine learning: from phenomena to black-box
  attacks using adversarial samples.
\newblock 2016{\natexlab{a}}.
\newblock URL \url{https://arxiv.org/abs/1605.07277}.

\bibitem[Papernot et~al.(2016{\natexlab{b}})Papernot, McDaniel, Jha,
  Fredrikson, Celik, and Swami]{sp2016:papernot}
Papernot, N., McDaniel, P., Jha, S., Fredrikson, M., Celik, Z.~B., and Swami,
  A.
\newblock The limitations of deep learning in adversarial settings.
\newblock In \emph{IEEE European Symposium on Security \& Privacy},
  2016{\natexlab{b}}.

\bibitem[Papernot et~al.(2016{\natexlab{c}})Papernot, McDaniel, Wu, Jha, and
  Swami]{papernot2016distillation}
Papernot, N., McDaniel, P., Wu, X., Jha, S., and Swami, A.
\newblock Distillation as a defense to adversarial perturbations against deep
  neural networks.
\newblock In \emph{Security and Privacy (SP), 2016 IEEE Symposium on}, pp.\
  582--597. IEEE, 2016{\natexlab{c}}.

\bibitem[Papernot et~al.(2017)Papernot, McDaniel, Goodfellow, Jha, Celik, and
  Swami]{papernot17}
Papernot, N., McDaniel, P., Goodfellow, I., Jha, S., Celik, Z.~B., and Swami,
  A.
\newblock Practical black-box attacks against machine learning.
\newblock In \emph{Proceedings of the 2017 ACM on Asia Conference on Computer
  and Communications Security}, ASIA CCS '17, pp.\  506--519, New York, NY,
  USA, 2017. ACM.
\newblock ISBN 978-1-4503-4944-4.
\newblock \doi{10.1145/3052973.3053009}.
\newblock URL \url{http://doi.acm.org/10.1145/3052973.3053009}.

\bibitem[Samangouei et~al.(2018)Samangouei, Kabkab, and
  Chellappa]{samangouei2018defensegan}
Samangouei, P., Kabkab, M., and Chellappa, R.
\newblock Defense-gan: Protecting classifiers against adversarial attacks using
  generative models.
\newblock \emph{International Conference on Learning Representations}, 2018.
\newblock URL \url{https://openreview.net/forum?id=BkJ3ibb0-}.
\newblock accepted as poster.

\bibitem[Sharif et~al.(2016)Sharif, Bhagavatula, Bauer, and
  Reiter]{ccs2016:sharif}
Sharif, M., Bhagavatula, S., Bauer, L., and Reiter, M.~K.
\newblock Accessorize to a crime: Real and stealthy attacks on state-of-the-art
  face recognition.
\newblock In \emph{Proceedings of the 2016 ACM SIGSAC Conference on Computer
  and Communications Security}, CCS '16, pp.\  1528--1540, New York, NY, USA,
  2016. ACM.
\newblock ISBN 978-1-4503-4139-4.
\newblock \doi{10.1145/2976749.2978392}.
\newblock URL \url{http://doi.acm.org/10.1145/2976749.2978392}.

\bibitem[Song et~al.(2018)Song, Kim, Nowozin, Ermon, and
  Kushman]{song2018pixeldefend}
Song, Y., Kim, T., Nowozin, S., Ermon, S., and Kushman, N.
\newblock Pixeldefend: Leveraging generative models to understand and defend
  against adversarial examples.
\newblock \emph{International Conference on Learning Representations}, 2018.
\newblock URL \url{https://openreview.net/forum?id=rJUYGxbCW}.
\newblock accepted as poster.

\bibitem[Szegedy et~al.(2013)Szegedy, Zaremba, Sutskever, Bruna, Erhan,
  Goodfellow, and Fergus]{szegedy-intriguing}
Szegedy, C., Zaremba, W., Sutskever, I., Bruna, J., Erhan, D., Goodfellow, I.,
  and Fergus, R.
\newblock Intriguing properties of neural networks.
\newblock 2013.
\newblock URL \url{https://arxiv.org/abs/1312.6199}.

\bibitem[Szegedy et~al.(2015)Szegedy, Vanhoucke, Ioffe, Shlens, and
  Wojna]{szegedy-inception}
Szegedy, C., Vanhoucke, V., Ioffe, S., Shlens, J., and Wojna, Z.
\newblock Rethinking the inception architecture for computer vision.
\newblock 2015.
\newblock URL \url{https://arxiv.org/abs/1512.00567}.

\bibitem[Xie et~al.(2018)Xie, Wang, Zhang, Ren, and Yuille]{xie2018mitigating}
Xie, C., Wang, J., Zhang, Z., Ren, Z., and Yuille, A.
\newblock Mitigating adversarial effects through randomization.
\newblock \emph{International Conference on Learning Representations}, 2018.
\newblock URL \url{https://openreview.net/forum?id=Sk9yuql0Z}.
\newblock accepted as poster.

\bibitem[Zantedeschi et~al.(2017)Zantedeschi, Nicolae, and
  Rawat]{zantedeschi2017efficient}
Zantedeschi, V., Nicolae, M.-I., and Rawat, A.
\newblock Efficient defenses against adversarial attacks.
\newblock \emph{arXiv preprint arXiv:1707.06728}, 2017.

\end{thebibliography}
